\documentclass[11pt, a4paper, copyright, gr]{google}

\usepackage[numbers]{natbib}

\usepackage{caption} 
\usepackage{textcomp} 
\usepackage{soul} 
\usepackage{mdframed}
\usepackage{needspace}
\usepackage{enumitem}
\usepackage{longtable}

\usepackage{float}
\usepackage{ltablex} 
\usepackage{array}
\usepackage{xurl}

\keepXColumns

\usepackage[utf8]{inputenc} 
\usepackage[T1]{fontenc}    
\usepackage{url}            
\usepackage{booktabs}       
\usepackage{amsfonts}       
\usepackage{nicefrac}       
\usepackage{microtype}      

\usepackage{amsmath}
\usepackage{algorithm}
\usepackage{algorithmicx}
\usepackage{algpseudocode}
\usepackage{multirow}
\usepackage{makecell}
\usepackage{graphicx}
\usepackage[normalem]{ulem} 
\usepackage[table]{xcolor}
\definecolor{bestgreen}{RGB}{226,239,218}
\usepackage{amssymb}
\usepackage{pifont}
\usepackage{subcaption}
\usepackage{hyperref}       

\usepackage[nameinlink, capitalise]{cleveref}

\title{Remote Sensing Vision-Language Models for Zero-Shot Multi-Retrieval}

\reportnumber{} 

\correspondingauthor{$\{$shekelto,genady$\}$@google.com}

\uselogo{}

\author[1]{Roy Hachnochi}
\author[1]{Bolous Jaber}
\author[2]{Vered Silverman}
\author[1]{Yotam Gigi}
\author[1]{George Leifman}
\author[1]{Tomer Shekel}
\author[1]{Genady Beryozkin}

\affil[1]{Google Research}
\affil[2]{Work done at Google via QualityAI}

\begin{abstract}

Contrastive Vision-Language Models (VLMs) have shown remarkable success in general visual understanding. However, reusing the same methods without adapting them to the unique properties of common Remote Sensing (RS) tasks undermines the real-world utility of the resulting models. 

Benchmarks serve as the design component that captures the essence of these unique RS tasks, and in addition to quantifying downstream utility, critically steer the model’s entire optimization trajectory.
Yet, while many remote sensing models use in-domain training data, they are rarely evaluated on the multi-retrieval nature of Earth observation, where an individual satellite tile contains multiple co-occurring geographic entities satisfying diverse queries, and a single localized query corresponds to numerous matching instances across a global archive. 

Benchmarks that are defined using classification and text-to-image retrieval tasks produce valuable metrics of a model's performance, but sole reliance on such benchmarks rewards models for macroscopic scene heuristics rather than granular visual analysis, steering them away from optimal real-world performance. Additionally, the contrastive, CLIP-style training objective also favors holistic, scene-level descriptions over localized, object-centric queries that are common in multi-retrieval benchmarks.

To optimize contrastive VLMs for better real-world performance, we establish a new benchmark suite that is closely aligned with common Earth observation tasks, formalizing evaluation as a continuous precision-ranked retrieval task. 
In tandem, we present a remote-sensing foundation VLM trained on a dataset exceeding 200 million examples. By combining Gemini-generated dense captions, landmark metadata, and additional localization cues, our setup is  optimized to recognize fine-grained spatial configurations across dense candidate pools. 
Empirical evaluations across both 224$\times$224 and native 512$\times$512 resolutions demonstrate that our model substantially outperforms existing state-of-the-art baselines on multi-retrieval benchmarks while matching or surpassing leading models on canonical classification and single-instance retrieval datasets.

\end{abstract}

\begin{document}

\maketitle

\section{Introduction}
Vision-Language Models (VLMs) have revolutionized open-world visual understanding, 
demonstrating unprecedented zero-shot classification, dense perception, and cross-modal 
retrieval capabilities~\cite{kuo2023mammut, tschannen2025siglip}. 
Foundational architectures such as CLIP~\cite{radford2021clip} established massive-scale contrastive language-image pre-training to map images and text into a shared geometric embedding space, pulling semantically aligned pairs together while repelling dissimilar distractors. 
Subsequent advances in general-domain representation learning have sought to unify holistic scene 
understanding with localized perception---exemplified by multi-teacher distillation frameworks 
like C-RADIO~\cite{ranzinger2026cradiov4}, which fuse the complementary representations of 
diverse foundation models.

Despite these rapid advancements, the real-world analytical utility of VLMs in the 
Remote Sensing (RS) domain remains severely constrained. 
Rather than a mere lag in model adaptation, this limitation stems from a fundamental, tripartite disconnect across: 
(1)~\textbf{dataset curation}, where paired text predominantly consists of coarse, holistic 
scene tags rather than granular, localized spatial descriptions; 
(2)~\textbf{training objectives}, which assume isolated 1-to-1 instance pairings and fail to 
supervise multi-target visual configurations; and 
(3)~\textbf{evaluation benchmarks}, which measure performance on rigid, single-instance 
matching tasks that do not reflect operational Earth observation workflows.

A primary bottleneck hindering the development of domain-specific RS foundation models has been the scarcity of diverse, high-quality, large-scale image-text datasets. With the challenge of sparse annotated data for this domain, initial attempts to train a dedicated RS-VLM focused on generating data for training using different strategies. 
Early attempts to adapt contrastive pre-training to remote sensing relied on template-based captions or small-scale datasets such as RSICD~\cite{lu2017exploring} and UCM-Captions~\cite{qu2016deep}, which now serve as standard benchmark datasets. Alternatively, GRAFT~\cite{mall2023graft} trains without text captions by matching satellite images to CLIP embeddings of geo-tagged ground-level photos from the same location, though ground photos are scarce in remote regions and often depict street-level details invisible from overhead. More recent efforts achieved scaling paired training data by tenfold. 
RemoteCLIP~\cite{liu2024remoteclip} converted existing object bounding boxes and masks into textual captions, SkyScript~\cite{wang2024skyscript} aligned OpenStreetMap semantic tags with satellite tiles to yield 2.6 million semantically diverse pairs, while RS5M and GeoRSCLIP~\cite{zhang2024rs5m} scaled pre-training data to 5 million pairs by combining web scraping with automated caption generation. 
Further improvement was achieved by DGTRS-CLIP~\cite{chen2025dgtrsd} using 2M image-text pairs focusing on caption length diversity, while extending the model's capacity to process lengthy queries. 
Unlike approaches focused on scaling pre-training datasets, MPS-CLIP~\citep{li2026mpsclip} uses a frozen CLIP backbone with parameter-efficient tuning and subimage perspectives. It employs a "Gated Global Attention" adapter to adjust the frozen CLIP representations for global context without requiring keyword inputs. Concurrently, LLM-extracted keywords guide SamGeo in producing semantic subimage views, which a Multi-Perspective Representation module then aggregates for fine-grained matching.

While early models focused on dataset scale, more recent models invested in high-quality captions. HQRS-CLIP~\cite{he2025enhancing} demonstrated comparable results to GeoRSCLIP~\cite{zhang2024rs5m} by using approximately 210K images with 1.26M–1.3M image-text pairs produced via multi-perspective MLLM/LLM generation.
FarSLIP~\cite{li2025farslip} preserves semantic coherence via patch-to-patch self-distillation on RS5M before aligning cropped object \texttt{[CLS]} tokens with category templates on MGRS-200k ($\sim$200k images, $>$1M boxes) using multi-positive contrastive learning. However, relying on discrete class templates over isolated crops limits its utility for open-ended, relational multi-retrieval.
Most recently,~\citet{barzilai2025recipe} and ~\citet{EarthAI2025} introduced a data curation framework that couples overhead tiles with Google Maps landmark metadata, using multimodal LLMs (Gemini) to generate rich, context-aware descriptions alongside filtered web-scale imagery (RS-Landmarks and RS-WebLI).
While these efforts demonstrate that rich metadata and automated captioning enhance zero-shot transfer, the descriptive granularity required to capture subtle global-scale spatial dynamics still falls short of capturing the fine-grained visual features required for actionable geospatial intelligence.

The performance of the aforementioned models was primarily evaluated on zero-shot transfer capabilities across a suite of established scene classification and retrieval benchmarks.
Scene classification benchmarks include Functional Map of the World (fMoW)~\citep{christie2018fmow}, RESISC45~\citep{Cheng2017RESISC45}, UCMerced (UCM)~\citep{yang2010bag}, and Aerial Image Dataset (AID)~\citep{xia2017aid}.
However, these classification benchmarks were designed for closed-set image classification, thus becoming problematic in a zero-shot setting when there is visual overlap between classes.
Multi-label benchmarks such as BigEarthNet-MM~\cite{sumbul2021bigearthnet} and high-resolution MLRSNet~\cite{qi2020mlrsnet} relax the single-label constraint, yet remain restricted to closed-set tag taxonomies (and low-resolution imagery in BigEarthNet-MM) rather than open-ended, compositional queries.

Traditional cross-modal retrieval benchmarks, such as RSICD~\citep{lu2017exploring}, RSITMD~\citep{yuan2022exploring}, and NWPU~\citep{Cheng2022NWPU}, formulate text--image retrieval as a rigid single-instance, binary matching task. In these settings, a broad scene-level caption (e.g., ``an aerial view of a commercial harbor'') is paired with a single ground-truth image. Consequently, models are rewarded for identifying macroscopic background textures and simplistic visual shortcuts (e.g., green equals farmland) rather than the granular visual evidence required for actionable intelligence. In practice, operational Earth observation tasks rarely revolve around broad scene classification; intelligence analysts instead demand targeted, localized, and intent-driven spatial queries, such as identifying ``two boats in close proximity'' or ``a construction site adjacent to a waterway''.

Current benchmarks fail to capture the multi-retrieval reality of Earth observation datasets, where a single localized textual query corresponds to dozens or thousands of relevant overhead image tiles across a global satellite archive. Conversely, a single high-resolution satellite tile contains multiple distinct geographic entities and spatial configurations, satisfying multiple independent analytical queries simultaneously.
By evaluating models through a rigid 1-to-1 binary recall metric, current benchmarks incentivize models to overfit to dominant scene cues, actively penalizing foundation models capable of granular spatial understanding and multi-instance retrieval.

In natural-image retrieval, ECCV Caption~\cite{chun2022eccv} mitigated 1-to-1 false negatives via machine-pooled, human-verified many-to-many associations and mAP ranking on MS-COCO, yet analogous large-scale evaluation remains absent in Earth observation. 
While recent concurrent work, such as SkySpector~\citep{ouerghemi2026skyspector}, evaluates multi-positive queries, it is confined to a closed pool of 972 images; in contrast, our benchmark suite spans diverse spatial scales up to 1M candidates.

To resolve these foundational gaps, we present a unified framework that directly bridges the 
divide between model pre-training and real-world multi-retrieval in Earth observation. Rather than evaluating standard models post-hoc, we design an end-to-end foundation model, 
\textbf{RS-SigLIP2MR}, optimized for multi-target spatial retrieval through dense 
Gemini captions, landmark metadata, localized object-crop supervision, and ranking-aware 
orthogonal regularization.

Our core contributions are threefold:
\begin{enumerate}
    \item \textbf{Novel Multi-Retrieval Benchmark Suite:} We establish a comprehensive 
    evaluation suite that consists of various benchmarks, designed specifically for Earth observation multi-retrieval tasks.
    \item \textbf{Multi-Target Foundation Architecture:} We introduce \textbf{RS-SigLIP2MR}, 
    trained on a multimodal corpus exceeding 200 million examples. We provide 
    both a standard 224$\times$224 model and, to our knowledge, the first dedicated contrastive dual-encoder remote sensing VLM fine-tuned natively at 512$\times$512 resolution.
    \item \textbf{State-of-the-Art Multi-Retrieval Performance:} We demonstrate that our architecture achieves decisive state-of-the-art gains across all multi-retrieval benchmarks, surpassing leading general-domain and remote sensing foundation models by substantial margins while matching or surpassing prior state-of-the-art performance on canonical classification and single-instance retrieval tasks.
\end{enumerate}

\section{Methods}
\label{sec:methods}

\subsection{Evaluation Benchmarks} 


The effectiveness of RS-VLMs is fundamentally tied to the quality of evaluation benchmarks. Common benchmarks traditionally focus on two main tasks: \textit{classification} and \textit{retrieval}. Each of these has its pitfalls and introduces biases to the model evaluation. \textit{Classification} benchmarks evaluate the model's ability to distinguish between a fixed set of classes, and attribute an image to \textbf{at most} one of them. \textit{Retrieval} benchmarks match each image to a single text caption (and vice versa), and evaluate the model's ability to match each caption to its \textbf{unique} image. Both of these evaluation methods and metrics have the problem of forcing an unnatural one-to-one match between texts and images. For example:
\begin{itemize}
  \item An image in a classification dataset might have both an \textit{``airplane''} and a \textit{``building''} in it. A classification metric that forces a single class assignment might incorrectly reduce metric points for this image.
  \item A retrieval dataset might have an image with a caption \textit{``a small island in a green pond''} and another similar image with a caption \textit{``a green pond surrounding an island''}, such that both images match each other's captions. A good model might mix these two images and still be correct, but will lose metric points by the forced one-to-one match in the retrieval metric.
\end{itemize}

\begin{figure}[tb]
    \centering
    \includegraphics[width=\linewidth]{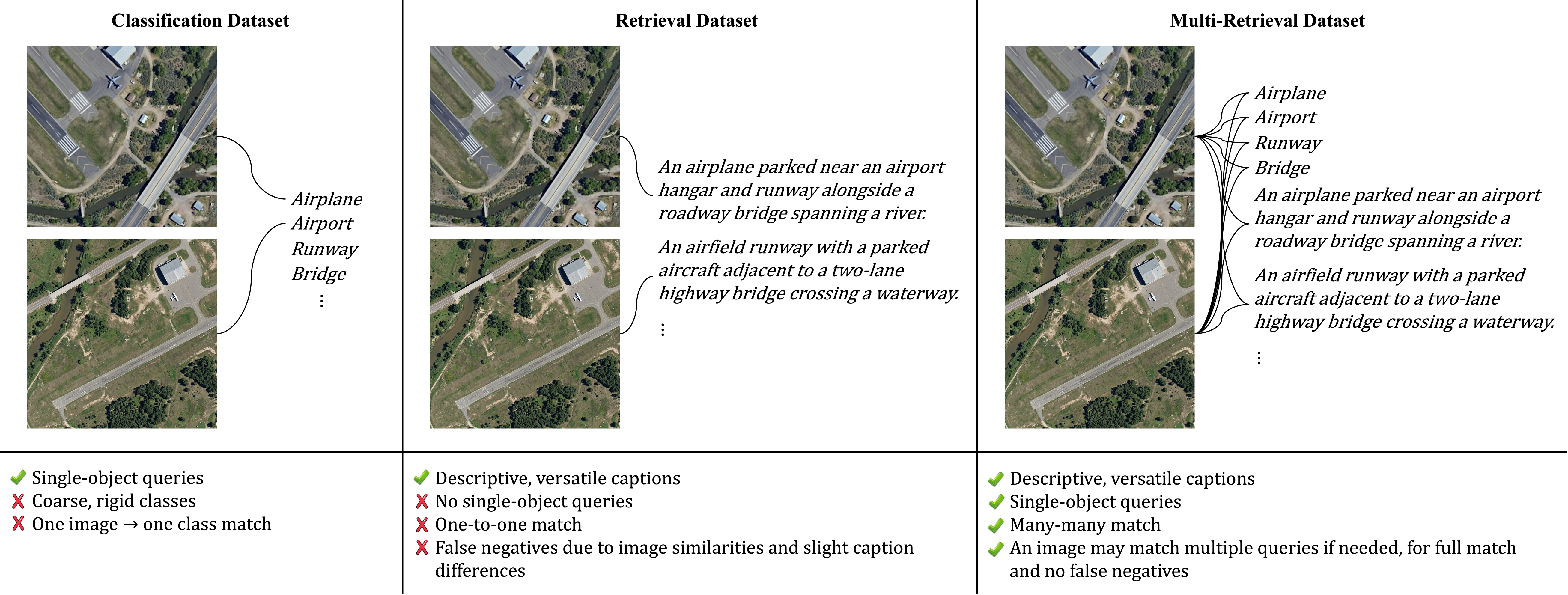}
    \caption{\textbf{Comparison of Classification, Retrieval, and Multi-Retrieval Datasets.} While Classification and Retrieval benchmarks force a one-to-one association, resulting in many shortcomings, the Multi-Retrieval dataset structure enables expressive, flexible many-to-many associations. Queries can span isolated semantic concepts to detailed compositional descriptions, and an individual image can correctly satisfy multiple distinct queries.}
    \label{fig:mr_comparison}
\end{figure}

However, common real-world usage of RS-VLMs often resembles a different scenario, where the user searches for a query, and the model retrieves \textbf{all} images that match this query. To better mirror this common use-case, we propose the \textbf{``Multi-Retrieval''} dataset structure, an accompanying metric, and several methodologies for its construction. \cref{fig:mr_comparison} highlights the shortcomings of common Classification and Retrieval datasets, and how the Multi-Retrieval datasets mitigate these issues.

\subsection{Dataset Architecture and Evaluation Method}
\label{subsec:multi_ret_arch}

A Multi-Retrieval benchmark dataset consists of three primary components: an \textit{image pool} $\mathcal{I} = \{i_1, i_2, \dots, i_n\}$, a \textit{query list} $\mathcal{Q} = \{q_1, q_2, \dots, q_m\}$, and a set of \textit{ground truth matches} for each query. For a given text query $q_j \in \mathcal{Q}$, its ground truth set $\text{GT}_j \subseteq \mathcal{I}$ specifies the set of images that correctly satisfy the statement or query. Note that ground truth sets are not mutually exclusive; an individual image may belong to the ground truth sets of multiple queries.

\begin{figure}[tb]
    \centering
    \includegraphics[width=\linewidth]{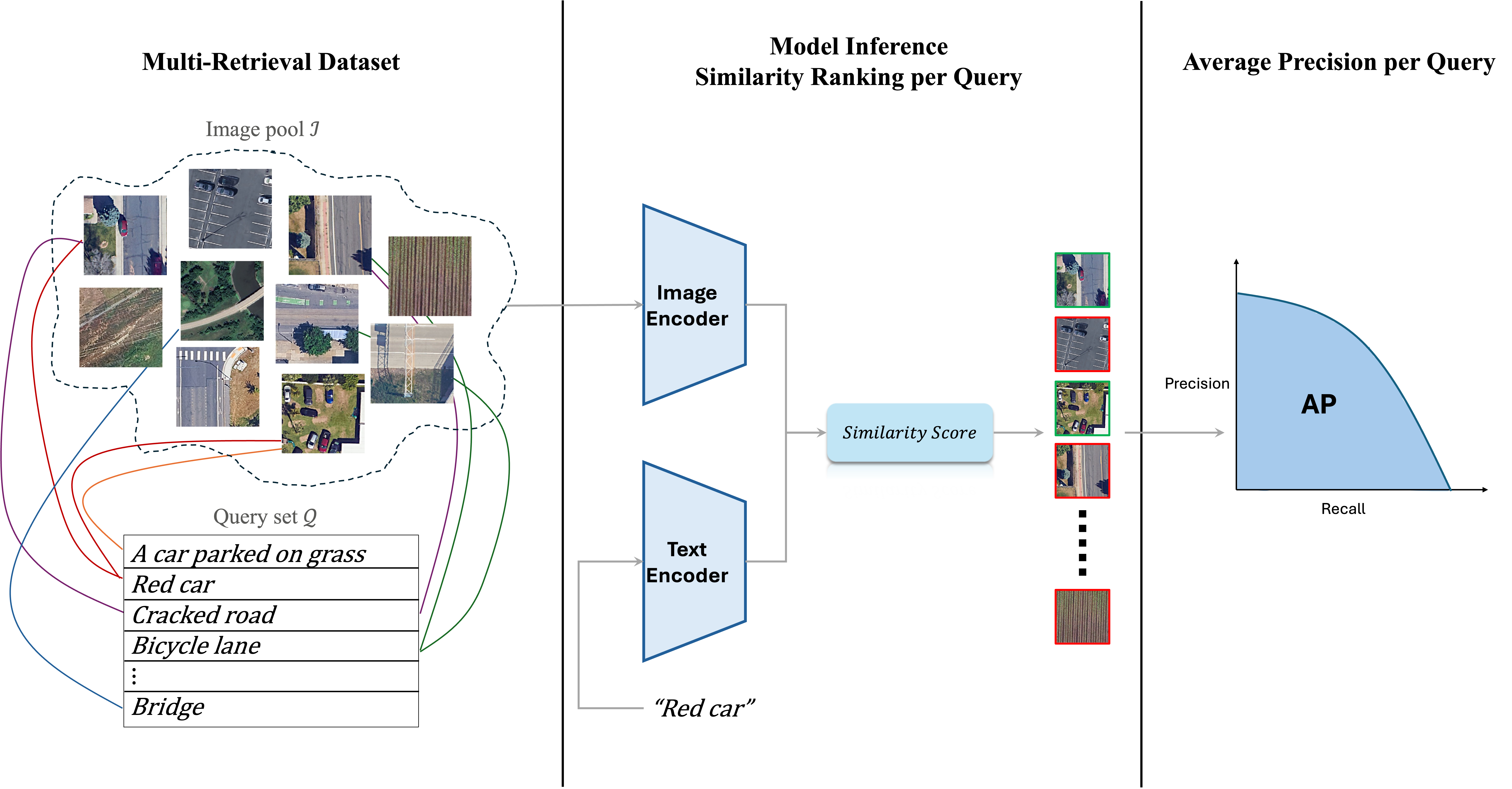}
    \caption{\textbf{Overview of the Multi-Retrieval Evaluation Pipeline.} Given an image pool $\mathcal{I}$, a query set $\mathcal{Q}$, and query-specific ground-truth sets $\text{GT}_j \subseteq \mathcal{I}$, a VLM computes pairwise similarity scores to rank the entire image corpus in descending order for each query. Rather than relying on rigid classification cutoffs, evaluation is framed as a continuous ranking task where performance is measured by Average Precision ($\text{AP}_j$) computed across the retrieved sequence, inherently balancing precision and recall without arbitrary thresholds.}
    \label{fig:eval_method}
\end{figure}

VLMs compute multi-modal text and image embeddings, which are most commonly used to calculate a similarity score per image vs. the text query. Our main insight is that this use-case may, and should, be formalized as a \textbf{ranking} task. We therefore opt to rank the entire image pool by their similarity scores with the query embedding, and calculate the \textit{mean average-precision (\textbf{mAP})} score as a metric. \cref{fig:eval_method} depicts the evaluation process of VLMs using the Multi-Retrieval benchmarks.

The Average Precision ($\text{AP}$) for a single query $q_j$ evaluates the quality of the ranked list over all recall levels, defined as:

\begin{equation}
\text{AP}_j = \frac{1}{|\text{GT}_j|} \sum_{r=1}^{n} \mathbb{I}_j(r) \cdot P_j(r), \qquad P_j(r) = \frac{1}{r} \sum_{i=1}^{r} \mathbb{I}_j(i)
\end{equation}
where $n$ is the size of the image pool, $|\text{GT}_j|$ is the size of the ground truth set for query $q_j$, $\mathbb{I}_j(r) \in \{0, 1\}$ is a relevance indicator returning $1$ if the item at rank $r$ is relevant ($i_r \in \text{GT}_j$) and $0$ otherwise, and $P_j(r)$ is the precision of query $j$ evaluated at rank $r$.

In practice, we use a novel weighted and normalized version of AP, defined as:
\begin{equation}
\text{AP}_j = \frac{\sum_{r=1}^{n} \mathbb{I}_j(r) \cdot P_j(r)}{\sum_{i=1}^{k_j} \frac{i}{w(i)}}, \qquad P_j(r) = \frac{1}{w(r)} \sum_{i=1}^{r} \mathbb{I}_j(i)
\end{equation}
where $k_j = |\text{GT}_j|$. We set $w(r) = \sqrt{r}$ for sub-linear discounting, giving more weight to the tail against standard weighting ($w(r) = r$). This weighting method is slightly more forgiving to lower-ranked items, which aids in comparing models that differ at the tail.

Finally, the \textit{mean average-precision} is defined as:
\begin{equation}
\text{mAP} = \frac{1}{|\mathcal{Q}|} \sum_{j=1}^{|\mathcal{Q}|} \text{AP}_j
\end{equation}

By accounting for precision at every position where a relevant item is retrieved, \textbf{mAP} provides a threshold-agnostic evaluation that inherently balances precision and recall, appropriately rewarding models that rank true positives higher in the retrieved sequence.

Note that following standard protocol across zero-shot vision-language literature~\citep{radford2021clip, liu2024remoteclip}, candidate ranking is performed via pairwise cosine similarity between text and image embeddings. 
We intentionally refrain from post-hoc manifold transformations (e.g., PCA whitening or learned re-rankers) to evaluate the native, unadulterated geometry of the zero-shot representation space directly.

\subsection{Evaluation Dataset Generation}
\label{subsec:multi_ret_generation}

Evaluating VLMs under the multi-retrieval paradigm requires benchmark datasets with diverse query complexities, spatial scales, and ground-truth distributions. Because exhaustive annotation over large image corpora is prohibitively expensive, we employ three complementary dataset construction strategies: (1) \textit{Small-scale, exhaustively annotated regional datasets}, (2) \textit{A large-scale 1M-image pool constructed via model-bootstrapped pooling}, and (3) \textit{Derived multi-retrieval benchmarks transformed from existing object detection datasets}. We also report results on the concurrent SkySpector~\citep{ouerghemi2026skyspector} benchmark.

Using the procedures described below, we generated dataset variants at spatial resolutions of 224$\times$224 and 512$\times$512 pixels. For most datasets, the 512$\times$512 variants share identical center coordinates and Ground Sample Distance (GSD) with their 224$\times$224 counterparts, but provide broader spatial context. 
This expanded context can aid model predictions by supplying surrounding environmental cues and capturing larger-scale spatial structures. On the other hand, the additional visual complexity and extraneous objects may reduce the relative signal of the relevant object within the image, potentially degrading retrieval performance. Conversely, DOTA2, FAIR1M, and SkySpector maintain their context in 224$\times$224 vs. 512$\times$512 variants, only changing the effective GSD.

\cref{tab:datasets} summarizes the Multi-Retrieval benchmarks evaluated in this report.
Appendix \cref{sec:app_labeling} complements this section and details the exhaustive labeling process of the manually curated datasets. For examples of queries used across the datasets, in addition to positive rates, please refer to Appendix \cref{sec:app_queries}.

\begin{table}[ht]
\centering
\small
\caption{Summary of curated evaluation benchmarks across scales and derivation paradigms.}
\label{tab:datasets}
\resizebox{\linewidth}{!}{%
\begin{tabular}{cllcccc}
\toprule
\textbf{Type} & \textbf{Dataset} & \textbf{Region / Context} & \textbf{|\(\mathcal{I}\)|} & \textbf{|\(\mathcal{Q}\)|} & \textbf{GSD} & \textbf{Matches per Query} \\
\midrule
\multirow{9}{*}{\shortstack{Exhaustively Annotated \\ Multi-Regional}} 
 & East Hanover   & USA (Suburban)      & $\sim400$       & 16  & 100 cm          & 1--5 \\
 & Gary           & USA (Urban/Ind.)    & $\sim2,700$     & 26  & 50 cm           & 1--5 \\
 & Mraclin 100    & Europe (Rural)      & $\sim300$       & 12  & 100 cm          & 1--5 \\
 & Mraclin 30     & Europe (Rural)      & $\sim3,300$     & 13  & 30 cm           & 1--5 \\
 & Najarpur       & India (Rural)       & $\sim600$       & 11  & 100 cm          & 1--5 \\
 & Pleasantville  & USA (Suburban)      & $\sim1,100$     & 15  & 30 cm           & 1--5 \\
 & Parkland Hills & USA (Residential)   & $\sim3,500$     & 20  & 10 cm           & 115 avg. \\
 & Astoria        & USA (Urban)         & $\sim2,600$     & 14  & 10 cm           & 37 avg. \\
 & Shibuya        & Japan (Urban)       & $\sim3,500$     & 13  & 30 cm           & 58 avg. \\
\midrule
 Large-Scale Pool via & \multirow{2}{*}{US Cover 1M} & \multirow{2}{*}{USA} & \multirow{2}{*}{$\sim1M$} & \multirow{2}{*}{33} & \multirow{2}{*}{10, 30, 100 cm} & \multirow{2}{*}{453 avg.} \\
 Model Bootstrapping & & & & & & \\
\midrule
\multirow{3}{*}{Derived} 
 & DOTA2~\cite{dotav2}          & Global              & $\sim20,000$    & 16  & 10 cm - 4.5 m   & $\sim1,800$ avg. \\
 & FAIR1M~\cite{sun2021fair1m}         & Global              & $\sim8,100$     & 37  & 30-80 cm        & $\sim660$ avg. \\
 & CTUS           & USA                 & $\sim10,000$    & 7   & 10, 30, 50 cm   & $\sim600$ avg. \\
\midrule
 Manually Annotated  & SkySpector~\citep{ouerghemi2026skyspector}     & Global              & $\sim1,000$     & 412 & 10 cm - 30 m    & 5 avg. \\
\bottomrule
\end{tabular}%
}
\end{table}

\subsubsection{Exhaustively Annotated Regional Benchmarks}
\label{subsec:small_scale_datasets}

To establish high-fidelity ground truth benchmarks with complete recall supervision, we curated several regional datasets spanning diverse geographic domains and GSD. For each region, image pools ranging from $\mathcal{O}(100)$ to $\mathcal{O}(1000)$ images were gathered over bounded geographic extents.

A set of fine-grained queries $\mathcal{Q}$ was formulated for each region, specifically targeting low-frequency semantic objects. To maximize recall fidelity, every image in the pool was exhaustively verified and annotated by at least 2 expert annotators. To enhance linguistic diversity, queries were augmented using textual synonyms.

\subsubsection{Large-Scale Image Pool via Model-Bootstrapped Pooling}
\label{subsec:large_scale_dataset}

To evaluate models under realistic retrieval regimes involving massive candidate sets, we constructed a large-scale evaluation pool of $|\mathcal{I}| \approx 1M$ images. However, exhaustive annotation of $1M$ candidates across dozens of queries is practically intractable. To address this challenge, we adopted a \textit{pooling and bootstrapping candidate selection protocol} ~\cite{voorhees2005trec,craswell2021trecdl}, a standard procedure in large-scale Information Retrieval benchmarks to make exhaustive human judging tractable over million-scale candidate corpora ~\cite{sanderson2010test,kuznetsova2020openimages}. The methodology proceeds as follows:
\begin{enumerate}
    \item \textbf{Candidate Retrieval:} For a curated set of $\mathcal{Q}$ distinct semantic queries, we execute zero-shot retrieval across an ensemble of $N$ diverse baseline models.
    \item \textbf{Top-\(k\) Union Pool:} For each query $q_j$ and model $\mathcal{M}_i$, we extract the top-$k$ ranked candidates from each evaluated model and form the union set:
    \begin{equation}
    \mathcal{U}_j = \bigcup_{i=1}^{N} \text{Top}_k\left(\mathcal{M}_i, q_j\right)
    \end{equation}
    \item \textbf{Manual Adjudication:} Expert annotators label only the candidate set $\mathcal{U}_j$, determining the positive ground-truth subset $\text{GT}_j \subseteq \mathcal{U}_j$.
    \item \textbf{Post-Hoc Metric Evaluation:} Model rankings are evaluated against this adjudicated pseudo-ground-truth.
\end{enumerate}
We select $k=30$ throughout our experiments, to balance annotation tractability and metric accuracy. For simplicity, and to avoid scarcity of available imagery for specific GSDs, we evaluate the models on datasets combined for all available GSDs, rather than evaluating separately.

While some remote sensing applications require exhaustive discovery of all target instances, evaluating the top-$k$ images closely mirrors interactive, human-in-the-loop workflows, where users input a query and often review only the top-$k$ images, disregarding the ranking order of the tail. Furthermore, this framework seamlessly accommodates the evaluation of newly introduced models by simply augmenting the existing union candidate pool $\mathcal{U}_j$ with their top-$k$ retrieved predictions, and manually labeling them. Notably, this pooling strategy represents a deliberate trade-off: it drastically reduces annotation overhead to make large-scale curation feasible, but shifts the evaluation focus toward precision over recall by restricting adjudication to the high-confidence head of the ranked list. While this approach contains inherent trade-offs and might present bias, Appendix \cref{subsec:head_truncation_tradeoff} provides additional insight and analysis on the validity of this head-truncation bootstrapping protocol.

\subsubsection{Derived Datasets via Spatial Windowing}
\label{subsec:derived_datasets}

To leverage rich existing bounding-box annotations from existing object detection (OD) benchmarks while preserving genuine multi-retrieval characteristics, we developed an automated spatial windowing transformation pipeline.

\begin{figure}[tb]
    \centering
    \includegraphics[width=0.7\linewidth]{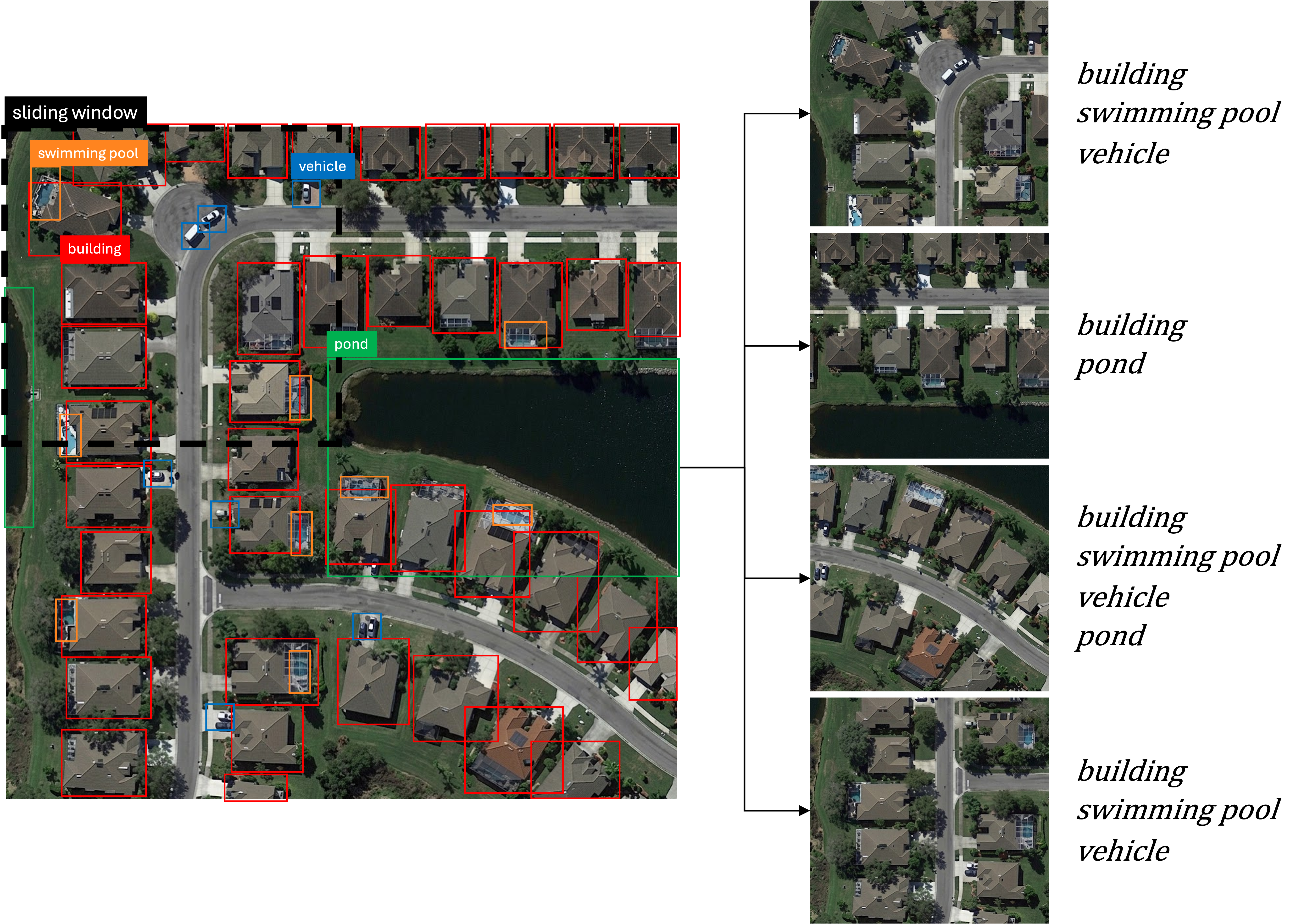}
    \caption{\textbf{Derived Multi-Retrieval Generation via Spatial Windowing.} A uniform sliding window partitions high-resolution scenes into image crops; objects whose bounding boxes are fully enclosed or satisfy a minimum spatial overlap threshold are attributed to the crop, and the resulting mapping is inverted to form multi-match query sets.}
\label{fig:derived_method}
\end{figure}

Directly cropping localized bounding boxes around individual objects produces single-object, centered isolated targets, biased towards a specific class, effectively reducing back to a \textit{classification} dataset. Instead, we generate the multi-retrieval benchmark by applying a uniform sliding window across source imagery, formalized in \cref{alg:spatial_windowing}. This sliding-window formulation, illustrated in \cref{fig:derived_method}, ensures that individual image crops frequently contain multiple co-occurring categories, strictly preserving the multi-label, multi-match dynamics essential for multi-retrieval evaluation. Crops may also turn out empty, which enriches the dataset with additional distractors.

Through this method, we transformed DOTA2~\citep{dotav2}, 
FAIR1M~\citep{sun2021fair1m}, and CTUS, a newly constructed dataset based on cellular tower locations provided by HIFLD~\citep{hifld_cellular_towers_2025}, and restructured them as Multi-Retrieval benchmarks. 
We set $\tau=10\text{ px}$ to exclude truncated objects; for CTUS we extract constant-GSD windows with $w=h=s=224$ and $w=h=s=512$, whereas for DOTA2 and FAIR1M we extract windows with $w=h=s=512$ and down-sample them to $224\times 224$ for the 224 benchmark, preserving spatial context.

\begin{algorithm}[tb]
\caption{Multi-Retrieval Dataset Construction via Spatial Windowing}
\label{alg:spatial_windowing}
\begin{algorithmic}[1]
\Require Annotated dataset $\mathcal{S} = \{(I_i, \mathcal{O}_i)\}$, window dimensions $w \times h$, stride $s$, threshold $\tau$.
\Ensure Image crop pool $\mathcal{I}$, Query set $\mathcal{Q}$, Ground-truth mappings $\{\text{GT}_q\}_{q \in \mathcal{Q}}$.

\State Initialize $\mathcal{I} \gets \emptyset$, $\mathcal{Q} \gets \emptyset$, $\text{GT}_q \gets \emptyset \quad \forall q$

\For{each annotated image $(I_i, \mathcal{O}_i) \in \mathcal{S}$}
    \State Extract crop bounding boxes and cropped images $\mathcal{C} = \{(b_c, I_c)\}$ from $I_i$ with dimensions $w \times h$ and stride $s$
    \State $\mathcal{I} \gets \mathcal{I} \cup \{I_c \mid (b_c, I_c) \in \mathcal{C}\}$
    
    \For{each crop $(b_c, I_c) \in \mathcal{C}$}
        \For{each object annotation $(b_k, t_k) \in \mathcal{O}_i$}
            \State Let $(w_{\cap}, h_{\cap})$ be the width and height of $b_k \cap b_c$
            \If{$w_{\cap} \ge \tau \ \textbf{and} \ h_{\cap} \ge \tau$}
                \State $\mathcal{Q} \gets \mathcal{Q} \cup \{t_k\}$
                \State $\text{GT}_{t_k} \gets \text{GT}_{t_k} \cup \{I_c\}$
            \EndIf
        \EndFor
    \EndFor
\EndFor

\State \Return $\mathcal{I}, \mathcal{Q}, \{\text{GT}_q\}_{q \in \mathcal{Q}}$
\end{algorithmic}
\end{algorithm}

\subsection{Training Datasets and Training Recipe} 
To enhance the model's multi-retrieval capabilities, we adopt the training dataset generation recipe introduced by~\cite{EarthAI2025}, improving upon it by incorporating targeted focus datasets. Similar to~\cite{EarthAI2025}, the core of our training corpus comprises three main datasets: RS-Landmarks, RS-WebLI, and RS-Global.
The RS-Landmarks dataset consists of satellite and aerial imagery aligned with landmarks extracted from Google Maps. We input this contextual information alongside the image into the Gemini model~\cite{gemini31flashimage}, utilizing a tailored prompt to generate descriptive captions.
The RS-WebLI dataset, along with corresponding alt-text from the WebLI dataset~\citep{chen2022pali}, was constructed by filtering the extensive WebLI corpus with specialized aerial and overhead image classifiers.
The RS-Global dataset features aerial and satellite images collected at native sensor resolutions ranging from 10 cm to 50 cm. This dataset provides nearly comprehensive global land coverage, with a distribution skewed toward areas of human activity, such as urban centers and agricultural regions. Each image is paired with a textual description generated by Gemini.
Crucially, we upgraded the data quality and class distribution of these collections beyond the original release in~\cite{EarthAI2025}. 
All candidate tiles underwent stringent artifact filtering and unsupervised visual clustering to prevent over-represented landscapes from dominating the training distribution. 
In parallel, captions were re-generated with deeper contextual metadata and filtered against hallucination and ambiguity to ensure faithful spatial grounding.

To enrich these three foundational datasets with highly focused data, we employ two complementary approaches. First, we leverage multiple proprietary Google object detection datasets, comprising  hundreds of thousands of aerial images annotated specifically for remote sensing tasks. 
This collection includes approximately 4 million labeled instances across more than 40 object categories, such as airplanes, baseball fields, boats, bridges, crosswalks, cell towers, and oil and gas wells. To generate image-caption pairs, we extract crops centered on the detection bounding boxes and feed them, along with their corresponding object categories, into Gemini using a custom prompt to produce detailed descriptions.
In the second approach, we utilize public catalogs to identify specific geographic locations featuring points of interest such as roads, airports, harbors, and rivers. We then acquire aerial images centered around these locations and annotate them using Gemini.

Our {\bf RS-SigLIP2MR} model was trained by fine-tuning the SigLIP2~\citep{tschannen2025siglip} checkpoint on the above dataset corpus in two steps.
We start the training on the three foundational datasets: RS-Landmarks, RS-WebLI, and RS-Global and then continue fine-tuning with the additional focused data.
All the above datasets were generated in two versions: 512$\times$512 and 224$\times$224 image sizes.
Based on the above image sizes, we train two variants of RS-SigLIP2MR:
\vspace{-0.1in}
\begin{itemize}
  \item RS-SigLIP2MR-224: Vision tower So400M variant of ViT with patch size 14 
  \item RS-SigLIP2MR-512: Vision tower So400M variant of ViT with patch size 16
\end{itemize}
\vspace{-0.1in}
\noindent
The text tower for both variants is So400M and the embedding size is 1152.
During fine-tuning, in addition to the contrastive SigLIP loss, we also employ the following two losses:
(1) \textit{Matryoshka Representation Learning (MRL) loss~\cite{kusupati2022matryoshka}}: a training technique that embeds data at multiple granularities within a single vector, allowing dynamic truncation of the embedding size to save memory and speed up searches without retraining.
(2) \textit{Global Orthogonal Regularization (GOR) loss~\cite{zhang2017learning}}: a regularization technique designed to ensure that learned representations (in our case, text/image embeddings) are maximally and uniformly distributed across the embedding space.
Incorporating GOR loss ensures that the high-dimensional latent space ($d=1152$) remains isotropic and uniform, mitigating anisotropic ``cone effects'' and preventing noise accumulation in the embedding tail from distorting cosine ranking across massive candidate galleries.

\section{Experimental Results}

\subsection{Zero-Shot Multi-Retrieval}
\label{subsec:multi_ret_results}
While traditional benchmarks assume an artificial 1-to-1 correspondence between text queries and images, real-world Earth observation workflows require locating all candidate scenes that satisfy complex spatial or semantic criteria. 
We evaluate our approach on the Multi-Retrieval (MR) benchmarks summarized in~\cref{tab:datasets}, across both 224$\times$224 and 512$\times$512 input resolutions. For brevity and visual clarity, we report the micro-averaged mAP across all multi-regional datasets.

\begin{figure}[h!]
    \centering
    \includegraphics[width=\linewidth]{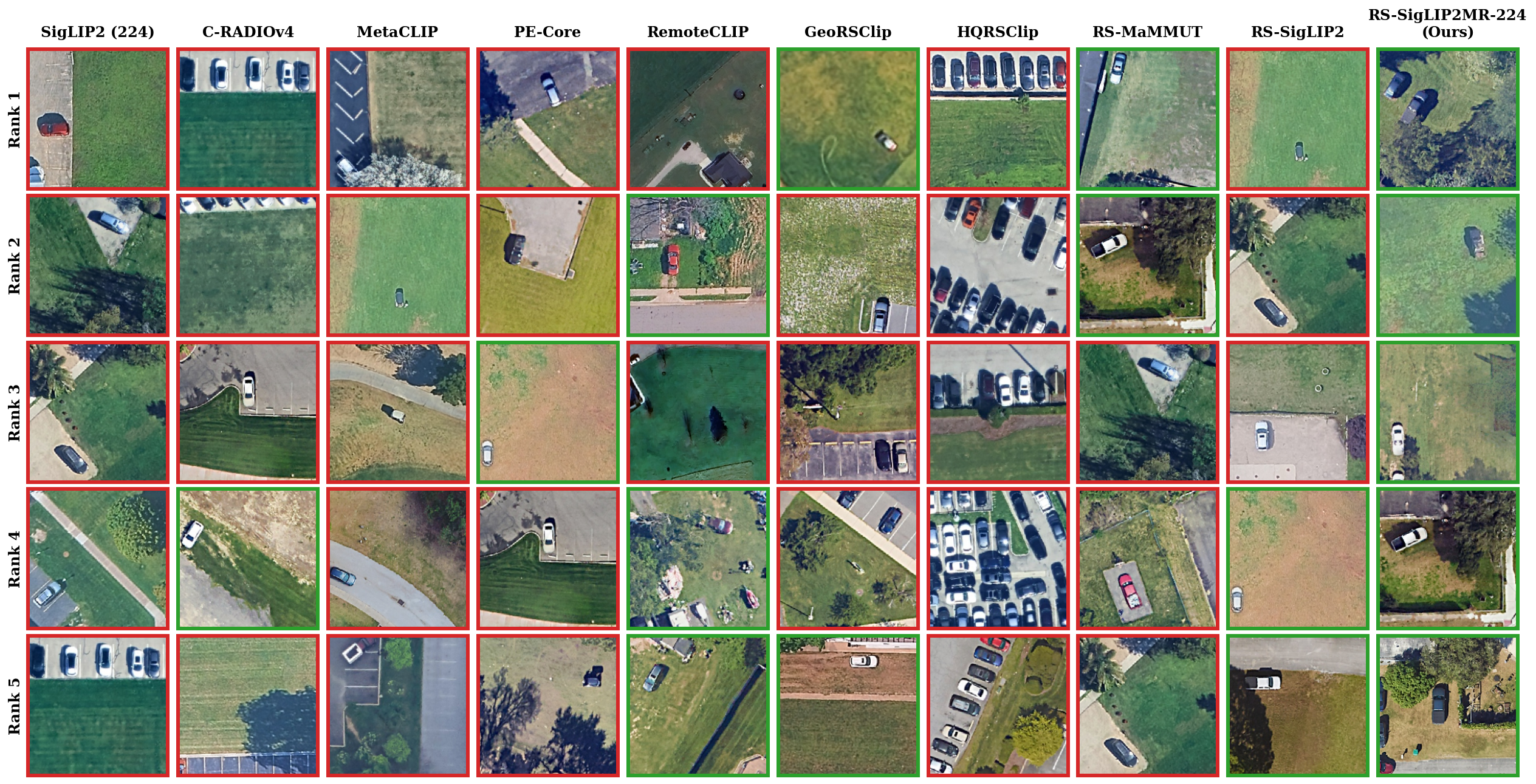}
    \caption{\textbf{Qualitative Comparison of Baselines on an Example Query.} We present the \textit{top-5} ranks of representative baselines on the query \textit{``car parked on grass''} from the \textit{US Cover 1M (224)} dataset. Positive images are marked with a green border, and negative images with red. Competing baselines exhibit a ``bag-of-words'' failure, whereas our model captures the semantic and spatial relations specified in the query.}
    \label{fig:results_comparison}
\end{figure}

As shown in~\cref{tab:multi_retrieval_combined}(a), at 224$\times$224 resolution \textbf{RS-SigLIP2MR-224} demonstrates a decisive advantage over all evaluated baselines, including general-purpose VLMs (e.g., PE-Core, C-RADIOv4, MetaCLIP2), dedicated remote sensing models (e.g., RemoteCLIP, GeoRSCLIP, HQRS-CLIP), and the base RS-SigLIP2. 
The performance gap is especially pronounced on benchmarks that require recognizing small, localized geographic features, where our model outperforms all competing models by a wide margin. 
This substantial boost directly validates our training strategy of incorporating object detection crops and localized text supervision. 
Furthermore, our model maintains a clear lead on regional scene matching and broader geographic queries. 
Crucially, on the large-scale manually labeled benchmark, our model maintains robust retrieval precision even in the presence of massive negative distractor sets, where competing models suffer substantial degradation. Specific mean Average Precision (mAP) scores are presented in~\cref{tab:multi_retrieval_combined}(a). \cref{fig:results_comparison} qualitatively compares retrieval rankings across representative baselines for a challenging query: \textit{``car parked on grass''}. While competing baseline models exhibit a classic ``bag-of-words'' failure mode—retrieving images containing disjointed instances of \textit{car} and \textit{grass}—our model successfully reasons over the spatial and semantic relationships specified in the query. Notably, our model maintains this precision even when retrieving from a highly challenging distractor pool of 1M images, demonstrating robust, fine-grained retrieval capabilities at scale.

To analyze how our architecture scales to higher input resolution and expanded spatial coverage, we evaluate \textbf{RS-SigLIP2MR-512} on 512$\times$512 imagery in~\cref{tab:multi_retrieval_combined}(b).
As detailed in~\cref{subsec:multi_ret_generation}, for our constant-GSD benchmarks (Multi-Regional, US Cover 1M, and CTUS), the 512$\times$512 construction expands the geographic footprint by approximately $5.2 \times$, whereas DOTA2, FAIR1M, and SkySpector preserve the scene extent at finer effective pixel resolution.
Consequently, on constant-GSD benchmarks, scaling to 512 induces a natural domain-dependent trade-off: for macro-scale categories and extensive cadastre, the broader spatial context provides vital topological grounding, whereas for localized discrete targets (e.g., cell towers in CTUS), the expanded field of view dilutes target signal within increased background clutter. Meanwhile, on fixed-extent benchmarks (DOTA2, FAIR1M, SkySpector), higher native resolution consistently improves fine-grained recognition. Additional results and discussion are provided in Appendix~\cref{sec:app_quantitative}.

Existing baselines span fixed 224-resolution encoders (including prior remote sensing VLMs), resolution-agnostic architectures (e.g., C-RADIOv4), and native 512-resolution general-domain models (e.g., SigLIP2 512), and we compare against all three categories.
Crucially, \textbf{RS-SigLIP2MR-512} consistently outperforms every competing model across all 512$\times$512 benchmarks, establishing a substantial margin over both high-resolution/resolution-agnostic general-domain VLMs and 224-resolution remote sensing baselines.

\begin{table}[h!]
    \centering
    \small
    \setlength{\tabcolsep}{7pt} 
    \caption{\textbf{Multi-Retrieval Results.} Mean Average Precision (mAP) across different image resolutions, showing that our models significantly outperform competing baselines. ($^\circ$Not a pure zero-shot evaluation for RS-SigLIP2MR; see text for details.)
    }
    \label{tab:multi_retrieval_combined}
    
    \begin{subtable}{\textwidth}
        \centering
        \caption{Multi Retrieval on 224$\times$224 imagery.}
        \resizebox{\linewidth}{!}{%
        \begin{tabular}{lcccccc}
            \hline
            \multirow{2}{*}{\textbf{Model}} & \multicolumn{3}{c}{\textbf{Manually Labeled}} & \multicolumn{3}{c}{\textbf{Derived from:}} \\
            \cline{2-7}
             & \textbf{Multi-Regional} & \textbf{US Cover 1M} & \textbf{SkySpector~\cite{ouerghemi2026skyspector}} & \textbf{CTUS$^\circ$} & \textbf{DOTA2~\cite{dotav2}} & \textbf{FAIR1M~\cite{sun2021fair1m}} \\
            \hline
            SigLIP2 (224)\citep{tschannen2025siglip} & 58.1 & 50.7 & 47.6 & 36.4 & 58.9 & 45.0 \\ 
            C-RADIOv4~\citep{ranzinger2026cradiov4}  & 60.1 & 62.1 & 49.6 & 34.8 & 61.1 & 45.9 \\ 
            MetaCLIP2~\cite{chuang2026meta}           & 52.4 & 43.5 & 43.9 & 35.3 & 55.3 & 42.9 \\ 
            PE-Core~\citep{bolya2025perception}      & 61.9 & 54.3 & 53.2 & 38.6 & 64.9 & 46.0 \\ 
            RemoteCLIP~\citep{liu2024remoteclip}     & 39.1 & 28.2 & 36.7 & 35.3 & 66.8 & 43.3 \\ 
            GeoRSCLIP~\citep{zhang2024rs5m}          & 54.6 & 42.8 & 41.6 & 33.0 & 52.5 & 38.7 \\ 
            HQRS-CLIP~\cite{he2025enhancing}         & 36.9 & 18.9 & 35.0 & 27.1 & 53.9 & 39.7 \\ 
            RS-MaMMUT\cite{EarthAI2025}              & 60.8 & 57.7 & 45.4 & 33.7 & 53.8 & 40.7 \\ 
            RS-SigLIP2~\cite{EarthAI2025}            & 60.4 & 42.1 & 46.4 & 34.3 & 56.2 & 43.7 \\ 
            \midrule
            RS-SigLIP2MR-224                     & \textbf{71.0} & \textbf{68.7} & \textbf{53.9} & \textbf{44.4} & \textbf{69.9} & \textbf{48.1} \\ 
            \hline
        \end{tabular}%
        }
    \end{subtable}
      
    \vspace{12pt} 
    
    \begin{subtable}{\textwidth}
        \centering
        \caption{Multi Retrieval on 512$\times$512 imagery.}
        \resizebox{\linewidth}{!}{%
        \begin{tabular}{lcccccc}
            \hline
            \multirow{2}{*}{\textbf{Model}} & \multicolumn{3}{c}{\textbf{Manually Labeled}} & \multicolumn{3}{c}{\textbf{Derived from:}} \\
            \cline{2-7}
             & \textbf{Multi-Regional} & \textbf{US Cover 1M} & \textbf{SkySpector~\cite{ouerghemi2026skyspector}} & \textbf{CTUS$^\circ$} & \textbf{DOTA2~\cite{dotav2}} & \textbf{FAIR1M~\cite{sun2021fair1m}} \\
            \hline
            SigLIP2 (512)\citep{tschannen2025siglip} & 49.9 & 53.7 & 49.8 & 35.6 & 59.5 & 46.4 \\ 
            C-RADIOv4~\citep{ranzinger2026cradiov4}  & 49.5 & 60.1 & 50.9 & 34.4 & 62.6 & 46.8 \\ 
            MetaCLIP2~\cite{chuang2026meta}           & 42.5 & 38.6 & 44.5 & 32.1 & 56.1 & 42.8 \\ 
            PE-Core~\citep{bolya2025perception}      & 53.6 & 55.1 & 53.5 & 35.2 & 65.7 & 46.7 \\ 
            RemoteCLIP~\citep{liu2024remoteclip}     & 27.9 & 20.8 & 31.3 & 25.8 & 64.1 & 37.6 \\ 
            RS-MaMMUT (224)\cite{EarthAI2025}        & 45.8 & 43.4 & 45.4 & 30.5 & 53.8 & 40.7 \\ 
            RS-SigLIP2 (224)~\cite{EarthAI2025}      & 47.0 & 39.0 & 46.1 & 31.6 & 56.2 & 43.7 \\ 
            \midrule
            RS-SigLIP2MR-224                     & 57.4 & 61.8 & 54.1 & 37.5 & 69.9 & 48.1 \\ 
            RS-SigLIP2MR-512                     & \textbf{60.1} & \textbf{68.0} & \textbf{55.4} & \textbf{38.5} & \textbf{70.9} & \textbf{49.2} \\ 
            \hline
        \end{tabular}%
        }
    \end{subtable}
   
\end{table}

Note that while the CTUS evaluation split was strictly held out from training, images from its disjoint training split were included in the fine-tuning of RS-SigLIP2MR. We therefore mark CTUS$^\circ$ in~\cref{tab:multi_retrieval_combined} to indicate that it represents an in-domain (non-pure zero-shot) evaluation for our models, whereas all baselines and all remaining Multi-Retrieval benchmarks are evaluated strictly zero-shot.


\subsection{Zero-Shot Classification and Retrieval}
To confirm that tuning our architecture for dense, multi-target retrieval does not degrade its fundamental cross-modal representations, we first benchmark \textbf{RS-SigLIP2MR} on canonical remote sensing zero-shot classification and traditional 1-to-1 text-to-image retrieval tasks.
\begin{table}[h!]
    \centering
    \small
    \setlength{\tabcolsep}{5pt}
    \caption{\textbf{Zero-Shot Classification Accuracy.} Top-1 accuracy across standard scene classification benchmarks. 
    Baseline numbers are taken from their respective papers.
    \textit{*Note: SkyScript evaluation was originally performed on an in-domain test set, rendering it a strictly easier task not directly comparable to zero-shot generalized architectures.
    The RemoteCLIP model was trained on images included in AID, making the evaluation not zero-shot.}}
    
    \label{tab:classification_results}
    \begin{tabular}{@{}lccccc@{}}
        \toprule 
        \textbf{Model} & \textbf{fMoW~\citep{christie2018fmow}} & \textbf{SkyScript~\citep{wang2024skyscript}} & \textbf{RESISC45~\citep{Cheng2017RESISC45}} & \textbf{UCM~\citep{yang2010bag}} & \textbf{AID~\citep{xia2017aid}} \\ 
        \midrule
        SkyScript~\citep{wang2024skyscript} & 28.04 & 70.89* & 70.94 & -- & -- \\
        RS-CLIP~\citep{li2023rsclip} & -- & 68.84 & 71.35 & 74.28 & 70.51 \\
        GeoRSCLIP~\citep{zhang2024rs5m} & -- & -- & 71.89 & -- & 76.33 \\
        RemoteCLIP~\citep{liu2024remoteclip} & -- & -- & 79.84 & -- & 91.30* \\
        MaMMUT~\citep{kuo2023mammut}  & 37.58 & 58.66 & 66.93 & 76.52 & 71.46 \\
        SigLIP2~\citep{tschannen2025siglip} & 41.25 & 65.94 & 72.40 & 81.43 & 75.56 \\
        RS-MaMMUT~\cite{EarthAI2025} & 47.24 & 69.46 & 72.31 & 80.29 & 71.96 \\
        RS-SigLIP2~\cite{EarthAI2025} & 48.13 & 68.13 & 80.13 & 84.86 & \textbf{78.26} \\ 
        \midrule
        RS-SigLIP2MR-224 & \textbf{50.62} & \textbf{70.74} & \textbf{80.36} & \textbf{88.48} & 75.96 \\ 
        \bottomrule
    \end{tabular}
\end{table}

\paragraph{Zero-Shot Classification.}
\cref{tab:classification_results} summarizes zero-shot classification performance across five standard remote sensing benchmarks: 
fMoW~\citep{christie2018fmow}, SkyScript~\citep{wang2024skyscript}, RESISC45~\citep{Cheng2017RESISC45}, UCM~\citep{yang2010bag}, and AID~\citep{xia2017aid}. 
Following standard evaluation conventions, models are prompted with generic class templates, i.e., \textit{``An aerial image of $<$class\_name$>$''}, and evaluated via nearest-neighbor matching. 

Overall, \textbf{RS-SigLIP2MR-224} achieves the highest zero-shot accuracy on four of the five benchmarks, outperforming both specialized remote sensing foundation models and general-domain baselines. Notably, our model surpasses the strong RS-SigLIP2 and RS-MaMMUT baselines on complex, temporally and geographically diverse datasets such as fMoW and SkyScript, establishes a clear lead on UCM, and achieves top accuracy on RESISC45. On AID, it remains competitive—modestly trailing RS-SigLIP2 and GeoRSCLIP while slightly outperforming general-domain SigLIP2 and MaMMUT. These trends demonstrate that optimizing for multi-retrieval does not induce catastrophic forgetting of broad semantic concepts; rather, learning from localized crops preserves—and in most cases enhances—category-level scene discrimination. Specific accuracy figures are detailed in~\cref{tab:classification_results}.

\paragraph{Zero-Shot Text-to-Image Retrieval.}
\cref{tab:retrieval_results} reports the average Top-1, 5, and 10 retrieval recall across four widely adopted benchmarks: RSICD~\citep{lu2017exploring}, UCM-Cap~\citep{qu2016deep}, RSITMD~\citep{yuan2022exploring}, and NWPU~\citep{Cheng2022NWPU}.
Across all four benchmarks, \textbf{RS-SigLIP2MR-224} consistently outperforms the strong RS-SigLIP2 baseline as well as all earlier specialized remote sensing models and general-domain VLMs, establishing leading zero-shot recall on every dataset. Improvements over RS-SigLIP2 are especially pronounced on RSICD, RSITMD, and UCM-Cap, with clear gains on NWPU as well. These results confirm that optimizing for localized multi-retrieval simultaneously strengthens global cross-modal alignment on conventional single-pair retrieval tasks. Specific recall metrics are provided in~\cref{tab:retrieval_results}.

\begin{table}[]
    \centering
    \small
    \setlength{\tabcolsep}{10pt} 
    \caption{\textbf{Zero-Shot Text-to-Image Retrieval.} Average Top-1/5/10 zero-shot retrieval recall across standard retrieval datasets. 
    Baseline numbers are taken from their respective papers.}
    \label{tab:retrieval_results}
    \begin{tabular}{@{}lcccc@{}}
        \toprule
        Model & RSICD~\citep{lu2017exploring} & UCM-Cap~\citep{qu2016deep} & RSITMD~\citep{yuan2022exploring} & NWPU~\citep{Cheng2022NWPU} \\
        \midrule
        PriorCLIP~\citep{pan2026priorclip} & 25.77 & -- & 39.85 & -- \\
        SkyScript~\citep{wang2024skyscript} & 19.97 & 59.33 & 30.58 & -- \\
        GeoRSCLIP~\citep{zhang2024rs5m} & 25.96 & -- & 38.02 & -- \\
        DGTRS-CLIP~\citep{chen2025dgtrsd} & 29.52 & -- & 39.57 & -- \\
        MaMMUT~\citep{kuo2023mammut} & 24.17 & 66.50 & 32.70 & 23.18 \\
        SigLIP2~\citep{tschannen2025siglip} & 27.93 & 68.52 & 36.70 & 25.14 \\
        RS-MaMMUT~\cite{EarthAI2025} & 33.59 & 71.79 & 42.58 & 32.28 \\
        RS-SigLIP2~\cite{EarthAI2025} & 37.64 & 75.33 & 47.26 & 37.74 \\ 
        \midrule
        RS-SigLIP2MR-224 & \textbf{39.69} & \textbf{76.66} & \textbf{48.66} & \textbf{38.41} \\ 

        \bottomrule
    \end{tabular}
\end{table}

\section{Conclusion and Future Work}
\label{sec:conclusion}

In this work, we addressed the fundamental limitation of conventional single-image cross-modal retrieval in remote sensing by formalizing the task of zero-shot multi-retrieval and establishing a comprehensive suite of Multi-Retrieval (MR) benchmarks—spanning urban-scale search, localized objects, fine-grained grounding, sparse infrastructure, and a million-scale distractor gallery. 
Alongside these benchmarks, we introduced \textbf{RS-SigLIP2MR}, a foundation model trained with localized crop supervision and multi-target objectives. 
Extensive evaluations across both 224$\times$224 and 512$\times$512 resolutions demonstrate that our model achieves substantial state-of-the-art gains on all MR benchmarks—particularly on small targets and massive candidate pools—while matching or surpassing leading baselines on canonical classification and traditional 1-to-1 retrieval tasks.

Moving forward, we plan to expand the capabilities of this framework by integrating a broader array of satellite and aerial data sources. While the current model relies primarily on high-resolution optical imagery, real-world Earth observation workflows benefit immensely from complementary sensor modalities. Incorporating Synthetic Aperture Radar (SAR), multispectral observations, and hyperspectral data into our multi-retrieval pre-training pipeline will enable more robust cross-sensor alignment and allow the model to operate effectively under diverse atmospheric and environmental conditions.

\bibliographystyle{plainnat}
\bibliography{main}

\clearpage
\appendix

\counterwithin{figure}{section}
\counterwithin{table}{section}

\begin{center}
  \vspace*{1em}
  {\LARGE \bfseries Remote Sensing Vision-Language Models for Zero-Shot Multi-Retrieval\\[0.5em] \Large Appendix \par}
  \vspace{2em}
\end{center}

\section{Annotation Protocol for Multi-Retrieval Datasets}
\label{sec:app_labeling}
Except for the derived benchmarks, the newly introduced Multi-Retrieval datasets were curated through an extensive manual labeling procedure detailed below. Annotations were conducted by a team of expert remote sensing imagery annotators who operated independently from the model development team to prevent evaluation bias.

\subsection{Multi-Regional Datasets}
\label{sec:app_small_scale_labeling}
The multi-regional benchmarks introduced in \cref{subsec:small_scale_datasets} feature highly specialized, region-specific queries with exhaustive binary ground truth (\textit{match}/\textit{no match}) across each regional image pool. After acquiring contiguous imagery that fully covers each designated region, we formulated candidate queries based on distinct geographical, structural, and infrastructural features present in that area.

We then performed exhaustive annotation over the entire pool to determine whether each image contained the target entity specified by the query. Each image was evaluated by two independent annotators; only instances achieving complete consensus were retained in the final ground truth. Furthermore, for quality control, an audit subset of 50 images per query (balanced to include both positive and negative candidates) was randomly sampled and inspected.

Finally, to increase the extent and diversity of used language, we duplicated queries with various synonyms and re-phrasings, while maintaining their GT sets identical. For example, \textit{"a building with distinct red roof"} $\rightarrow$ \textit{"a house featuring a prominent red roof"}, \textit{"housing project in development"} $\rightarrow$ \textit{"residential development under construction"}.

\subsection{Large-Scale Image Pool Benchmark}
\label{sec:app_large_scale_labeling}
Exhaustively annotating a pool of $1\text{M}$ images across 33 distinct queries is computationally and financially prohibitive (\cref{subsec:large_scale_dataset}). To make this tractable, we adopted a model-bootstrapped candidate generation strategy. For each evaluated model, we retrieved the top-$k$ predictions ($k = 30$) per query and constructed an aggregated candidate set consisting of the union of all top-$k$ retrievals across models.

Each image in the resulting candidate pool was independently reviewed by three annotators. Disagreements were adjudicated manually by a dedicated senior reviewer, who also inspected a randomly sampled validation subset of 100 images per query. In addition, we conducted blind manual spot-checks on a randomly chosen subset, with each sample independently cross-verified by another co-author.

\section{Examples of Queries Per Multi-Retrieval Dataset}
\label{sec:app_queries}
To supplement the dataset descriptions in the main text (\cref{subsec:multi_ret_generation}), \cref{tab:retrieval_dataset_queries} provides a representative sample of text queries evaluated across each Multi-Retrieval benchmark, along with image pool sizes and ground-truth positive counts. As illustrated, the queries capture a diverse spectrum of retrieval challenges, ranging from broad macroscopic features (e.g., \textit{crosswalk}, \textit{solar panel}) to fine-grained infrastructure subtypes (e.g., \textit{guyed cell tower}), relational context (e.g., \textit{swimming pool on a roof}, \textit{car parked on grass}), and rare targets with fewer than twenty matching instances (e.g., \textit{roller coaster}, \textit{airplane flying midair}). This diverse set of datasets and queries also reduces sampling-size bias by differing in the number of positive matches per query, allowing to validate models both on simple and difficult tasks.

\begin{table*}[p]
\centering
\footnotesize
\setlength{\tabcolsep}{4.5pt}
\renewcommand{\arraystretch}{0.90}
\setlength{\aboverulesep}{1.5pt}
\setlength{\belowrulesep}{1.8pt}
\captionof{table}{\textbf{Sample text queries from the Multi-Retrieval datasets}. Datasets with fully annotated evaluation pools indicate ``All'' in the tagged column, whereas US Cover 1M operates over an open million-image pool with query-specific rated sets.%
}
\label{tab:retrieval_dataset_queries}
\begin{tabularx}{\textwidth}{lc>{\raggedright\arraybackslash}Xrr}
\toprule
\textbf{Dataset} & \textbf{Image pool size} & \textbf{Query} & \textbf{\# Tagged} & \textbf{\# Positives} \\
\midrule
\multirow{12}{*}{Multi-Regional} 
  & \multirow{12}{*}{$\sim$2,000} 
  & \textit{a swimming pool surrounded by quonset huts} & \multirow{12}{*}{All} & 1 \\
  & & \textit{airplane flying} & & 1 \\
  & & \textit{a house featuring a prominent red roof} & & 3 \\
  & & \textit{solar panels on a large industrial building} & & 3 \\
  & & \textit{crosswalk with many people crossing} & & 3 \\
  & & \textit{a factory chimney with smoke} & & 7 \\
  & & \textit{soccer court} & & 7 \\
  & & \textit{rv parked in the home's driveway} & & 10 \\
  & & \textit{a house surrounded by gravel} & & 13 \\
  & & \textit{high voltage electrical powerlines} & & 18 \\
  & & \textit{building with an internal courtyard/lightwell} & & 62 \\
  & & \textit{stepping stones in a garden} & & 279 \\
\midrule
\multirow{8}{*}{US Cover 1M} 
  & \multirow{8}{*}{$\sim$1M} 
  & \textit{roller coaster} & 2,376 & 12 \\
  & & \textit{lifeguard stand} & 2,880 & 15 \\
  & & \textit{airplane flying midair} & 2,451 & 18 \\
  & & \textit{skybridge connecting two buildings} & 2,104 & 36 \\
  & & \textit{swimming pool on a roof} & 1,815 & 39 \\
  & & \textit{wind turbine} & 983 & 160 \\
  & & \textit{car parked on grass} & 2,046 & 307 \\
  & & \textit{crosswalk} & 9,740 & 1,461 \\
\midrule
\multirow{6}{*}{DOTA2} 
  & \multirow{6}{*}{$\sim$20K} 
  & \textit{container-crane} & \multirow{6}{*}{All} & 18 \\
  & & \textit{helicopter} & & 177 \\
  & & \textit{swimming-pool} & & 780 \\
  & & \textit{storage-tank} & & 1,202 \\
  & & \textit{harbor} & & 3,359 \\
  & & \textit{small-vehicle} & & 7,771 \\
\midrule
\multirow{6}{*}{FAIR1M} 
  & \multirow{6}{*}{$\sim$8K} 
  & \textit{c919} & \multirow{6}{*}{All} & 7 \\
  & & \textit{excavator} & & 86 \\
  & & \textit{boeing747} & & 159 \\
  & & \textit{tugboat} & & 360 \\
  & & \textit{intersection} & & 992 \\
  & & \textit{small car} & & 2,762 \\
\midrule
\multirow{6}{*}{SkySpector} 
  & \multirow{6}{*}{$\sim$1,000} 
  & \textit{A cluster of three vertical silos within a port} & \multirow{6}{*}{All} & 1 \\
  & & \textit{Bridge located south-west of stadium} & & 2 \\
  & & \textit{Surface mine with terraced open pits and haul roads} & & 4 \\
  & & \textit{A pond exhibiting a vivid green algal surface} & & 5 \\
  & & \textit{residential houses damaged by a hurricane} & & 16 \\
  & & \textit{Small parking lot with parked cars} & & 44 \\
\midrule
\multirow{7}{*}{CTUS} 
  & \multirow{7}{*}{$\sim$10K} 
  & \textit{cell tower on a silo} & \multirow{7}{*}{All} & 9 \\
  & & \textit{cell tower on a water tank} & & 78 \\
  & & \textit{utility pole cell tower} & & 167 \\
  & & \textit{monopole cell tower} & & 203 \\
  & & \textit{guyed cell tower} & & 431 \\
  & & \textit{lattice cell tower} & & 470 \\
  & & \textit{cell tower} & & 2,752 \\
\bottomrule
\end{tabularx}
\end{table*}

\section{Multi-Retrieval Multi-Regional Per-Dataset Results}
\label{sec:app_small_scale_metrics}

As described in \cref{subsec:multi_ret_results}, in the main paper we report micro-averaged metrics across the Multi-Retrieval regional datasets. For a detailed per-dataset breakdown across different imaging resolutions, we refer to \cref{tab:app_multi_retrieval_small_scale}. 

Across standard 224$\times$224 imagery (\cref{tab:multi_retrieval_224}), our proposed \mbox{RS-SigLIP2MR-224} model establishes a new state-of-the-art across all nine geographical regions, outperforming both general-domain vision models (SigLIP2, C-RADIOv4) and dedicated remote sensing baselines (RS-MaMMUT, RS-SigLIP2, GeoRSCLIP). Crucially, the model excels in ultra-high-resolution, dense urban scenes (Parkland Hills, Astoria, Shibuya), surpassing the strongest baselines by up to $+7.8$ AP points. This highlights the representation's robustness across diverse Ground Sample Distances (GSD, $10\,\text{cm}$ to $100\,\text{cm}$) and varied urban morphology worldwide.

Under expanded 512$\times$512 spatial context (\cref{tab:multi_retrieval_512}), standard 224-trained baselines degrade severely when evaluated on larger image tiles in dense urban scenes due to extraneous background clutter. In contrast, our dedicated \mbox{RS-SigLIP2MR-512} model effectively leverages the $4\times$ wider geographic extent: it achieves peak retrieval performance across complex landscapes.

\begin{table*}[t]
    \centering
    \small
    \caption{\textbf{Multi-Retrieval Multi-Regional Per-Dataset Results, 224 and 512.} Mean Average Precision across nine regional evaluation sites. The reported metrics on RS-SigLIP2MR models use pure zero-shot versions.}
    \label{tab:app_multi_retrieval_small_scale}
    
    \begin{subtable}{\textwidth}
        \centering
        \caption{Multi Retrieval on 224$\times$224 imagery.}
        \label{tab:multi_retrieval_224}
        \resizebox{\linewidth}{!}{%
        \begin{tabular}{lcccccccc}
            \toprule
            \textbf{Model} & 
            \makecell{\textbf{East}\\\textbf{Hanover}} & 
            \makecell{\textbf{Mraclin}\\\textbf{100}} & 
            \makecell{\textbf{Mraclin}\\\textbf{30}} & 
            \textbf{Najarpur} & 
            \textbf{Pleasantville} & 
            \makecell{\textbf{Parkland}\\\textbf{Hills}} & 
            \textbf{Astoria} & 
            \textbf{Shibuya} \\
            \midrule
            SigLIP2 (224)~\citep{tschannen2025siglip} & 58.5 & 71.7 & 61.7 & 58.1 & 69.5 & 45.6 & 44.7 & 45.7 \\
            C-RADIOv4~\citep{ranzinger2026cradiov4}   & 60.4 & 83.0 & 64.7 & 76.3 & 68.7 & 47.1 & 43.1 & 51.1 \\
            MetaCLIP2~\citep{chuang2026meta}           & 53.0 & 70.0 & 63.7 & 57.1 & 59.5 & 39.1 & 37.4 & 42.1 \\
            PE-Core~\citep{bolya2025perception}      & 61.6 & 80.1 & 77.2 & 67.6 & 65.8 & 53.0 & 48.3 & 50.8 \\
            RemoteCLIP~\citep{liu2024remoteclip}     & 30.8 & 47.1 & 50.6 & 62.6 & 41.5 & 29.3 & 30.4 & 30.4 \\
            GeoRSCLIP~\citep{zhang2024rs5m}          & 57.4 & 79.0 & 57.9 & 70.0 & 64.4 & 33.8 & 36.3 & 42.2 \\
            HQRS-CLIP~\citep{he2025enhancing}         & 35.2 & 38.1 & 45.4 & 49.7 & 45.6 & 31.7 & 30.0 & 24.7 \\
            RS-MaMMUT~\citep{EarthAI2025}              & 64.8 & 79.5 & 69.6 & 68.7 & 71.0 & 38.0 & 38.6 & 45.7 \\
            RS-SigLIP2~\citep{EarthAI2025}            & 61.3 & 77.8 & 66.3 & 80.2 & 61.8 & 45.9 & 42.3 & 47.8 \\
            \midrule
            RS-SigLIP2MR-224                & \textbf{72.3} & \textbf{90.6} & \textbf{78.7} & \textbf{85.3} & \textbf{78.0} & \textbf{60.3} & \textbf{51.6} & \textbf{58.9} \\
            \bottomrule
        \end{tabular}%
        }
    \end{subtable}
    
    \vspace{12pt}
    
    \begin{subtable}{\textwidth}
        \centering
        \caption{Multi Retrieval on 512$\times$512 imagery.}
        \label{tab:multi_retrieval_512}
        \resizebox{\linewidth}{!}{%
        \begin{tabular}{lcccccccc}
            \toprule
            \textbf{Model} & 
            \makecell{\textbf{East}\\\textbf{Hanover}} & 
            \makecell{\textbf{Mraclin}\\\textbf{100}} & 
            \makecell{\textbf{Mraclin}\\\textbf{30}} & 
            \textbf{Najarpur} & 
            \textbf{Pleasantville} & 
            \makecell{\textbf{Parkland}\\\textbf{Hills}} & 
            \textbf{Astoria} & 
            \textbf{Shibuya} \\
            \midrule
            SigLIP2 (512)~\citep{tschannen2025siglip} & 48.4 & 72.1 & 55.1 & 49.8 & 65.4 & 45.2 & 39.4 & 36.4 \\
            C-RADIOv4~\citep{ranzinger2026cradiov4}  & 48.7 & 68.7 & 53.9 & 53.0 & 61.5 & 44.3 & 42.5 & 39.6 \\
            MetaCLIP2~\citep{chuang2026meta}           & 49.7 & 53.3 & 45.9 & 39.0 & 53.4 & 39.8 & 32.9 & 30.4 \\
            PE-Core~\citep{bolya2025perception}      & 52.3 & 74.5 & 56.8 & 60.5 & 66.4 & 45.6 & 44.7 & 44.1 \\
            RS-MaMMUT (224)~\citep{EarthAI2025}        & 44.7 & 62.2 & 39.8 & 50.4 & 60.1 & 37.9 & 36.3 & 32.3 \\
            RS-SigLIP2 (224)~\citep{EarthAI2025}      & 46.8 & 68.1 & 35.1 & 55.1 & 58.7 & 41.7 & 37.1 & 35.7 \\
            \midrule
            RS-SigLIP2MR-224                & 54.2 & 73.9 & \textbf{66.2} & 58.8 & 73.8 & 50.0 & 42.1 & 43.9 \\
            RS-SigLIP2MR-512                & \textbf{56.7} & \textbf{91.8} & \textbf{66.2} & \textbf{66.6} & \textbf{74.7} & \textbf{53.0} & \textbf{45.0} & \textbf{51.6} \\
            \bottomrule
        \end{tabular}%
        }
    \end{subtable}
\end{table*}

\section{Additional Qualitative Results}
\label{sec:app_qualitative}

\begin{figure}[tb]
    \centering
    \includegraphics[width=\linewidth]{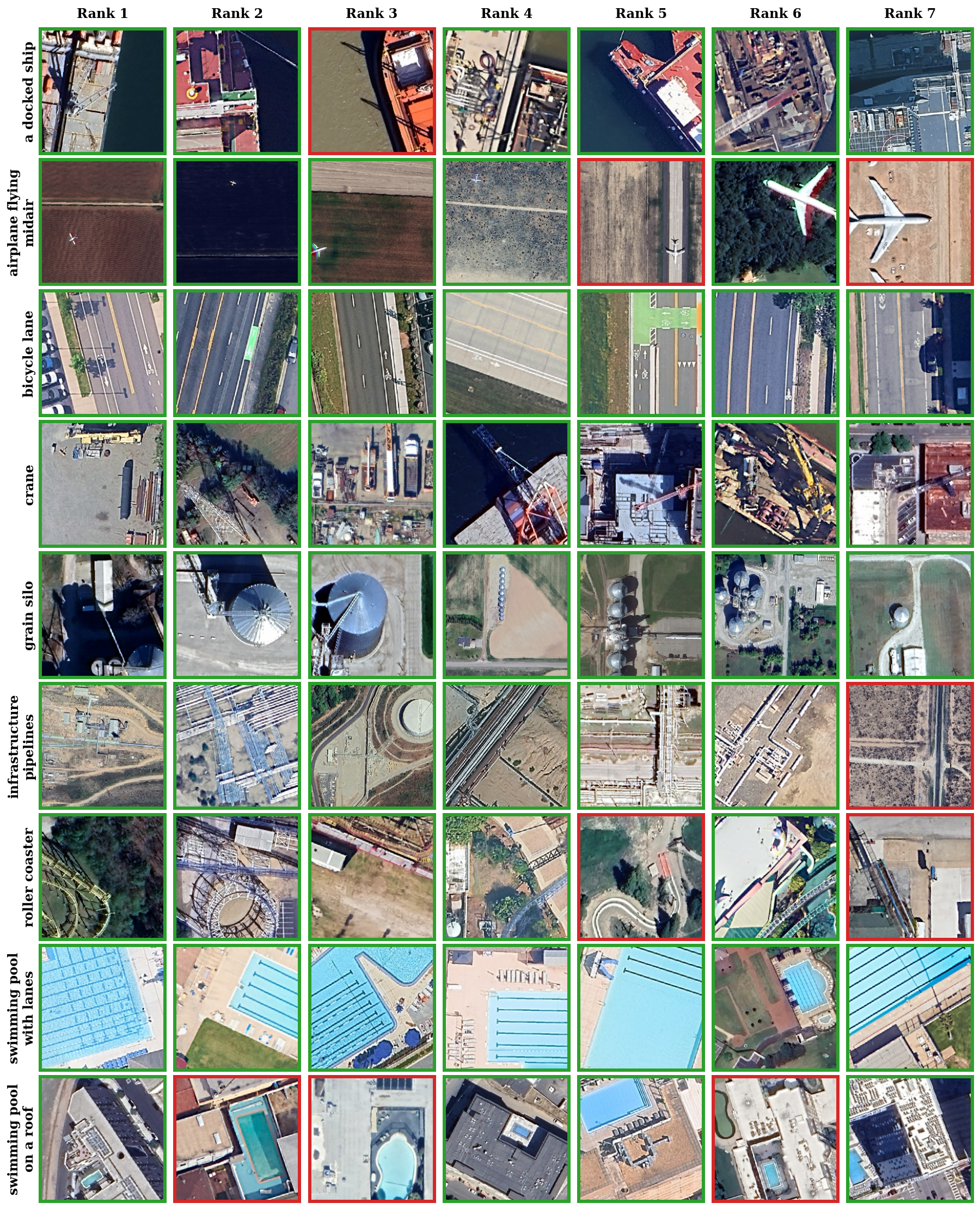}
    \caption{\textbf{Qualitative Results on \textit{US Cover 1M (224)} Dataset.} We present the \textit{top-7} ranks of our \textit{RS-SigLIP2MR-224} model on various representative queries from the \textit{US Cover 1M (224)} Dataset. Queries vary from simple objects, scarce/rare objects, and complex relation queries with multiple objects. Positive images are marked with a green border, and negative images with red. This figure also serves to show tagging examples from the dataset.}
    \label{fig:results_us_cover_1m}
\end{figure}

\begin{figure}[tb]
    \centering
    \includegraphics[width=\linewidth]{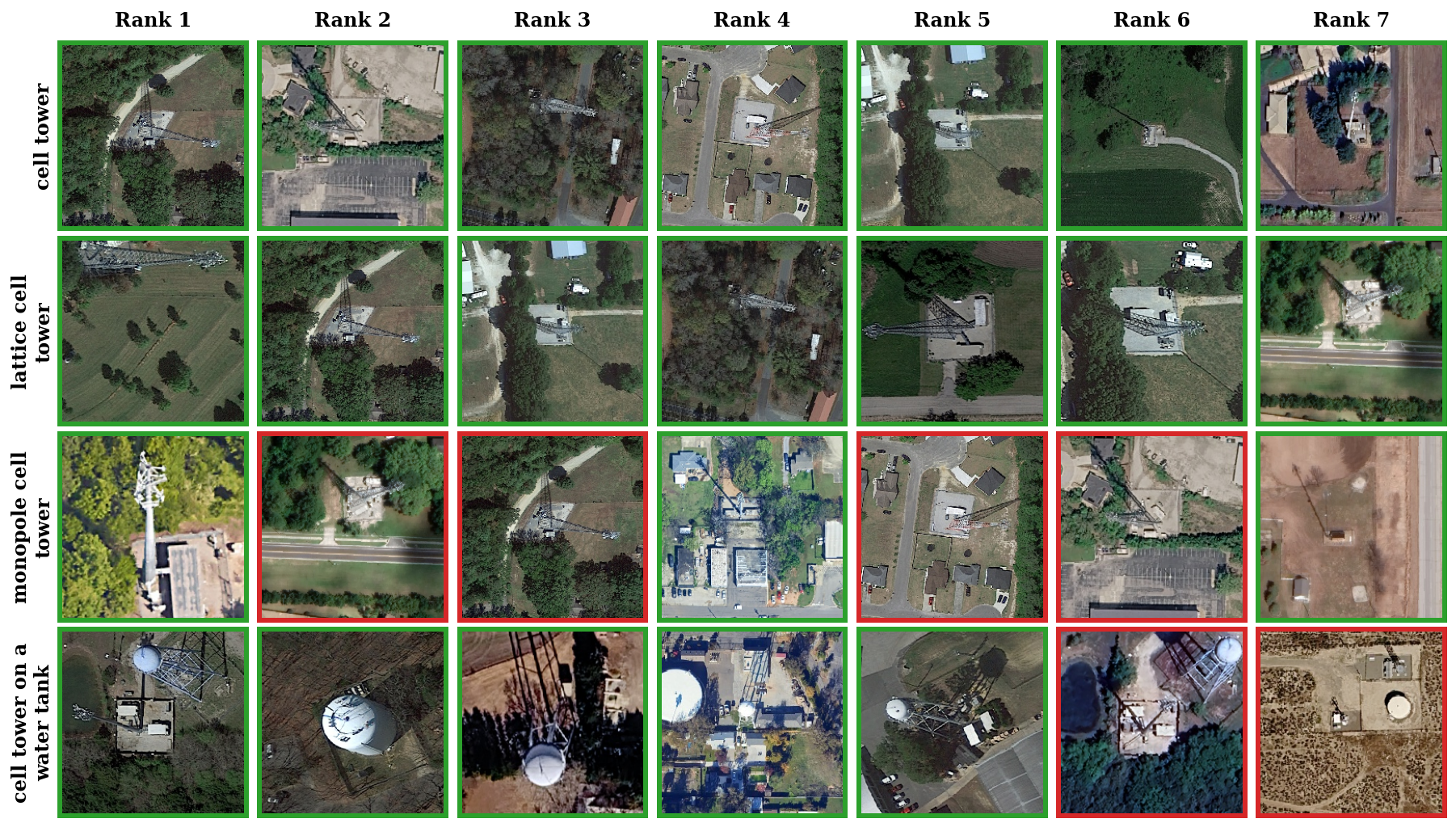}
    \caption{\textbf{Qualitative Results on \textit{CTUS (224)} Dataset.} We present the \textit{top-7} ranks of our \textit{RS-SigLIP2MR-224} model on various representative queries from the \textit{CTUS (224)} Dataset. The dataset contains a general \textit{cell tower} query to retrieve any cell tower type, and specific queries to distinguish between cell tower types. Positive images are marked with a green border, and negative images with red. This figure also serves to show tagging examples from the dataset.}
    \label{fig:results_ctus}
\end{figure}

\begin{figure}[tb]
    \centering
    \includegraphics[width=0.985\linewidth]{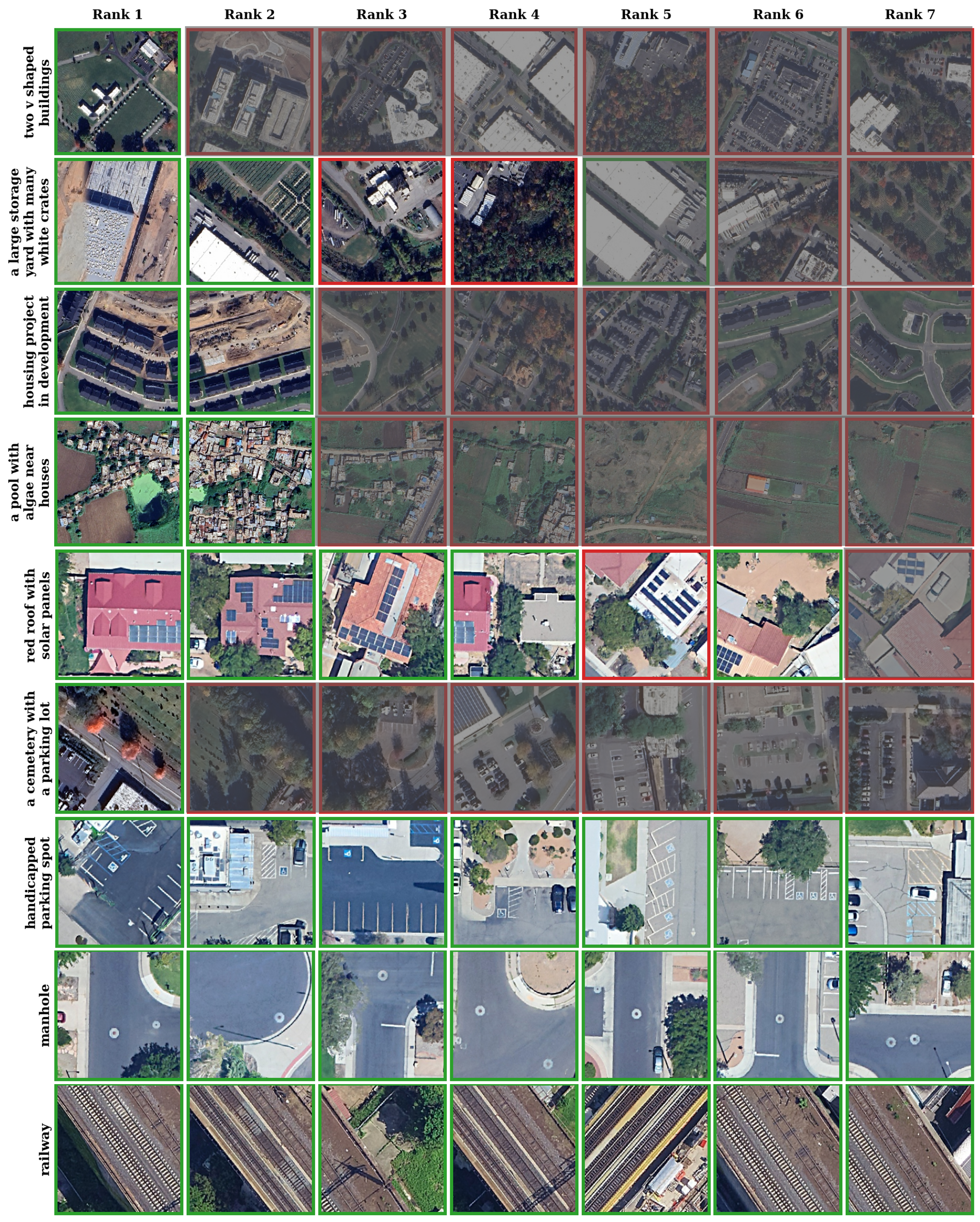}
    \caption{\textbf{Qualitative Results on \textit{Multi-Regional (224)} Datasets.} We present the \textit{top-7} ranks for our \textit{RS-SigLIP2MR-224} model across representative queries from the \textit{Multi-Regional (224)} datasets, which also illustrate sample annotations from our curation pipeline. Positive matches are denoted by green borders and false positives by red borders. For queries with $N < 7$ ground-truth matches in the pool, we gray out the retrieval ranks beyond $N$ to indicate the number of expected positive retrievals.
}
    \label{fig:results_multi_regional}
\end{figure}

~\cref{fig:results_us_cover_1m,fig:results_ctus,fig:results_multi_regional} showcase our models' ranking capabilities across a diverse range of queries from the Multi-Retrieval benchmarks. These figures illustrate the visual complexity and rigor of the evaluation sets by displaying both correct retrievals (indicated by green borders) and false positives (indicated by red borders). Across various queries of varying complexity, our models demonstrate robust retrieval performance. Even in cases where the models fail, the retrieved false positives exhibit strong semantic and visual relations to the query, pointing to understandable visual ambiguities rather than random failures.

\cref{fig:results_multi_regional} demonstrates our models' exceptional performance in "needle-in-a-haystack" retrieval scenarios. In this figure, retrieval ranks after the actual total number of ground-truth positives available in the search pool have been grayed out. For highly specific queries with as little as 1-4 positive targets among thousands of candidates (such as \textit{"two v-shaped buildings}" or \textit{"a cemetery with a parking lot"}), our models successfully retrieve and rank the true positive at Rank 1. This highlights the outstanding precision of our models for low-frequency target search in overhead imagery.

\section{Additional Quantitative Results}
\label{sec:app_quantitative}

\subsection{Mitigating Query Template Bias via Template Selection}
\label{sec:appendix_prefix_analysis}

Evaluating zero-shot VLMs with a single fixed prompt template (or bare queries alone) introduces a strong linguistic bias, as different models are pre-trained with distinct captioning styles and syntactic priors. Furthermore, because Multi-Retrieval queries range from single-word object tags to full relational phrases, any single static prefix inevitably penalizes models whose text encoders expect a different framing. To decouple visual retrieval capability from prompt sensitivity and reduce template bias, we evaluate each query across a bank of 13 diverse overhead, aerial, and satellite prefixes (including the bare query) and report the best-performing prefix per query in Table~\ref{tab:multi_retrieval_best_prefix}.

Two main conclusions emerge from this analysis. First, general-purpose and template-trained baselines rely heavily on prefix selection (preferring an added prefix on $66.1\%\text{--}83.4\%$ of queries) and achieve noticeably larger $\Delta\text{mAP}$ gains than remote-sensing adapted models, confirming that bare queries under-specify the overhead domain for non-specialized encoders. Second, our \textbf{RS-SigLIP2MR} models are substantially less sensitive to prompt phrasing---retaining the unmodified query far more frequently---while preserving overall state-of-the-art retrieval: at $224\times224$ only \text{PE-Core} slightly edges \textbf{RS-SigLIP2MR-224} on \textit{SkySpector} and \textit{FAIR1M}, whereas \textbf{RS-SigLIP2MR} leads decisively at $512\times512$.

\begin{table}[h!]
    \centering
    \small
    \setlength{\tabcolsep}{3.5pt}
    \caption{\textbf{Multi-Retrieval Results with Best-Prefix Selection.} Mean Average Precision (mAP) when each query selects its highest-scoring template from our 13-prefix bank, alongside the average gain over bare-query evaluation ($\Delta$\textbf{mAP}) and the share of queries selecting a non-empty prefix (\textbf{Alt-Pref.}). 
    }
    \label{tab:multi_retrieval_best_prefix}
    
    \begin{subtable}{\textwidth}
        \centering
        \caption{Multi Retrieval on 224$\times$224 imagery.}
        \begin{tabular}{lcccccccc}
            \hline
            \multirow{2}{*}{\textbf{Model}} & \multicolumn{3}{c}{\textbf{Manually Labeled}} & \multicolumn{3}{c}{\textbf{Derived from:}} & \multicolumn{2}{c}{\textbf{Prefix Stats}} \\
            \cline{2-9}
             & \textbf{Multi-Reg.} & \textbf{US Cover 1M} & \textbf{SkySpector} & \textbf{DOTA2} & \textbf{FAIR1M} & \textbf{CTUS} & $\Delta$\textbf{mAP} & \textbf{Alt-Pref.} \\
            \hline
            SigLIP2 (224)~\citep{tschannen2025siglip} & 63.6 & 61.9 & 56.2 & 64.4 & 47.9 & 38.2 & +4.6 & 76.4\% \\
            C-RADIOv4~\citep{ranzinger2026cradiov4}  & 64.5 & 67.4 & 55.8 & 66.7 & 46.9 & 37.1 & +3.7 & 66.1\% \\
            MetaCLIP2~\citep{chuang2026meta}           & 59.3 & 59.7 & 52.2 & 60.5 & 45.2 & 38.6 & +4.9 & 79.3\% \\
            PE-Core~\citep{bolya2025perception}      & 67.2 & 70.1 & \textbf{60.0} & 70.2 & \textbf{48.4} & 40.4 & +4.1 & 75.5\% \\
            RemoteCLIP~\citep{liu2024remoteclip}     & 43.3 & 53.7 & 41.3 & 69.5 & 45.3 & 38.4 & +3.0 & 78.8\% \\
            RS-MaMMUT~\citep{EarthAI2025}              & 64.5 & 58.8 & 50.3 & 55.3 & 41.1 & 34.6 & +2.2 & 67.2\% \\
            RS-SigLIP2~\citep{EarthAI2025}            & 63.0 & 54.4 & 50.4 & 56.2 & 44.2 & 34.6 & +1.7 & 50.6\% \\
            \midrule
            RS-SigLIP2MR-224                & \textbf{73.4} & \textbf{71.2} & 57.6 & \textbf{70.8} & 48.2 & \textbf{45.8} & +2.6 & 58.9\% \\
            \hline
        \end{tabular}
    \end{subtable}
    
    \vspace{12pt}
    
    \begin{subtable}{\textwidth}
        \centering
        \caption{Multi Retrieval on 512$\times$512 imagery.}
        \begin{tabular}{lcccccccc}
            \hline
            \multirow{2}{*}{\textbf{Model}} & \multicolumn{3}{c}{\textbf{Manually Labeled}} & \multicolumn{3}{c}{\textbf{Derived from:}} & \multicolumn{2}{c}{\textbf{Prefix Stats}} \\
            \cline{2-9}
             & \textbf{Multi-Reg.} & \textbf{US Cover 1M} & \textbf{SkySpector} & \textbf{DOTA2} & \textbf{FAIR1M} & \textbf{CTUS} & $\Delta$\textbf{mAP} & \textbf{Alt-Pref.} \\
            \hline
            SigLIP2 (512)~\citep{tschannen2025siglip} & 54.0 & 62.5 & 58.3 & 65.2 & 49.3 & 37.5 & +4.5 & 79.4\% \\
            C-RADIOv4~\citep{ranzinger2026cradiov4}  & 53.8 & 65.4 & 57.3 & 67.5 & 47.7 & 36.1 & +3.5 & 69.5\% \\
            MetaCLIP2~\citep{chuang2026meta}           & 47.2 & 54.9 & 53.0 & 60.9 & 45.2 & 34.9 & +4.5 & 83.4\% \\
            PE-Core~\citep{bolya2025perception}      & 57.3 & 64.9 & 60.1 & 70.6 & 49.0 & 37.1 & +3.7 & 77.5\% \\
            RemoteCLIP~\citep{liu2024remoteclip}     & 30.5 & 44.3 & 34.8 & 66.6 & 39.5 & 27.8 & +2.4 & 82.7\% \\
            RS-MaMMUT (224)~\citep{EarthAI2025}        & 48.8 & 49.6 & 50.2 & 55.4 & 41.1 & 31.1 & +1.9 & 69.3\% \\
            RS-SigLIP2 (224)~\citep{EarthAI2025}      & 49.4 & 47.4 & 50.2 & 56.2 & 44.2 & 32.0 & +1.3 & 51.4\% \\
            \midrule
            RS-SigLIP2MR-224                & 58.8 & 64.9 & 57.3 & 70.7 & 48.2 & \textbf{38.6} & +2.1 & 58.3\% \\
            RS-SigLIP2MR-512                & \textbf{63.2} & \textbf{69.9} & \textbf{61.3} & \textbf{74.9} & \textbf{50.5} & 37.4 & +2.9 & 64.4\% \\
            \hline
        \end{tabular}
    \end{subtable}
\end{table}

\subsection{Impact of Spatial Context and Pixel Resolution on Query Semantics}
\label{subsec:query_context_tradeoff}

To investigate how visual representations behave across varied search intents, \cref{tab:query_results} reports query-level retrieval precision across two distinct evaluation regimes: benchmarks with constant Ground Sample Distance (GSD), where $512\times512$ tiles capture a $4\times$ wider geographic footprint than $224\times224$ tiles (Multi-Regional, CTUS, and US Cover 1M), and benchmarks with fixed spatial extent, where both models observe the exact same geographic crop and $224\times224$ inputs are downsampled (DOTA2, FAIR1M, and SkySpector).

Under constant GSD, retrieval efficacy is governed by the alignment between contextual field of view and target scale. For expansive landscape features and site-level infrastructure, the wider spatial context of \mbox{RS-SigLIP2MR-512} yields substantial gains---such as \textit{a construction site} (+35.1 AP), \textit{a lake with brown water} (+30.0 AP), \textit{retail district} (+15.7 AP), and \textit{bicycle lane} (+5.2 AP)---where surrounding road networks and terrain help disambiguate the facility. Conversely, compact or localized targets exhibit steep drops under $4\times$ wider tiles, including \textit{two v shaped buildings} ($-85.1$ AP), \textit{vehicle-filled gravel lot} ($-46.9$ AP), \textit{cell tower on a water tank} ($-35.4$ AP), and \textit{tractor} ($-25.2$ AP), as expanded neighborhood coverage introduces competing distractors and severe target dilution.

Holding spatial context fixed isolates the effect of pixel fidelity from context expansion. Without added background clutter, dense small targets no longer suffer dilution, and \mbox{RS-SigLIP2MR-512} yields strong gains whenever downsampling blurs discriminative sub-meter structure or fine-grained subtypes---such as \textit{A zumwalt-class stealth destroyer warship seen from overhead} (+44.8 AP), \textit{Empty wastewater tank revealing internal scraper arms} (+24.9 AP), \textit{boeing747} (+9.9 AP), \textit{roundabout} (+9.7 AP), and \textit{helicopter} (+8.1 AP). Conversely, \mbox{RS-SigLIP2MR-224} remains stronger on coarse umbrella categories and holistic land-cover scenes, including \textit{An area of dense forest} ($-24.5$ AP), \textit{vehicle} ($-12.5$ AP), and \textit{ground-track-field} ($-7.8$ AP), where low-frequency layout and global texture suffice without fine detail. Together, these regimes show that higher pixel resolution consistently aids fine-grained discrimination within a fixed scene, whereas expanding geographic footprint is beneficial only when the broader context matches the semantic scale of the query.

\begin{table*}[t]
\centering
\small
\setlength{\tabcolsep}{3pt}
\caption{\textbf{Per-Query Multi-Retrieval Metrics}. Average Precision (\%) comparing \mbox{RS-SigLIP2MR-224} and \mbox{RS-SigLIP2MR-512} at their native resolutions (224$\times$224 vs.\ 512$\times$512). On Multi-Regional, CTUS, and US Cover 1M (top), both pools maintain constant GSD so 512$\times$512 tiles cover $4\times$ wider geographic context; on DOTA2, FAIR1M, and SkySpector (bottom), both models observe the same spatial footprint with 224$\times$224 downsampled.}
\label{tab:query_results}
\begin{tabular}{clcc}
\toprule
\makecell{\textbf{Dataset}} & \multicolumn{1}{c}{\textbf{Text Query}} & \makecell{\textbf{RS-SigLIP2MR}\\\textbf{-224}} & \makecell{\textbf{RS-SigLIP2MR}\\\textbf{-512}} \\
\midrule
\multicolumn{4}{l}{\textit{Constant GSD ($512\times512$ captures $4\times$ wider spatial context)}} \\
\midrule
\multirow{5}{*}{Multi-Regional} 
  & \textit{a construction site} & 57.7 & \textbf{92.8} \\
  & \textit{retail district} & 53.4 & \textbf{69.1} \\
  & \textit{a lake with brown water} & 17.2 & \textbf{47.2} \\
  & \textit{vehicle-filled gravel lot} & \textbf{86.2} & 39.3 \\
  & \textit{two v shaped buildings} & \textbf{100.0} & 14.9 \\
\midrule
\multirow{3}{*}{CTUS} 
  & \textit{cell tower} & \textbf{96.5} & 94.7 \\
  & \textit{lattice cell tower} & \textbf{50.4} & 45.5 \\
  & \textit{cell tower on a water tank} & \textbf{65.0} & 29.6 \\
\midrule
\multirow{4}{*}{US Cover 1M} 
  & \textit{a red car} & \textbf{100.0} & 92.8 \\
  & \textit{bicycle lane} & 94.8 & \textbf{100.0} \\
  & \textit{car parked on grass} & \textbf{65.5} & 50.3 \\
  & \textit{tractor} & \textbf{88.1} & 62.9 \\
\midrule
\multicolumn{4}{l}{\textit{Fixed Spatial Context ($224\times224$ downsampled from $512\times512$)}} \\
\midrule
\multirow{5}{*}{DOTA2} 
  & \textit{roundabout} & 58.5 & \textbf{68.2} \\
  & \textit{swimming-pool} & 61.5 & \textbf{70.8} \\
  & \textit{helicopter} & 53.4 & \textbf{61.5} \\
  & \textit{small-vehicle} & 79.4 & \textbf{80.7} \\
  & \textit{ground-track-field} & \textbf{48.6} & 40.8 \\
\midrule
\multirow{5}{*}{FAIR1M} 
  & \textit{bridge} & 39.7 & \textbf{50.9} \\
  & \textit{boeing747} & 49.8 & \textbf{59.7} \\
  & \textit{a350} & 31.2 & \textbf{41.0} \\
  & \textit{dump truck} & \textbf{50.7} & 43.0 \\
  & \textit{vehicle} & \textbf{66.0} & 53.5 \\
\midrule
\multirow{5}{*}{SkySpector} 
  & \textit{A zumwalt-class stealth destroyer warship seen from overhead} & 55.2 & \textbf{100.0} \\
  & \textit{Empty wastewater tank revealing internal scraper arms} & 22.0 & \textbf{46.9} \\
  & \textit{Small vehicle parked next to helicopter on ground} & 78.5 & \textbf{100.0} \\
  & \textit{Aerial image dominated by a river and its surrounding banks} & \textbf{59.6} & 49.2 \\
  & \textit{An area of dense forest} & \textbf{73.6} & 49.1 \\
\bottomrule
\end{tabular}
\end{table*}

\subsection{Head-Truncation Trade-offs.}
\label{subsec:head_truncation_tradeoff}
\begin{figure}[H]
    \centering
    \includegraphics[width=\linewidth]{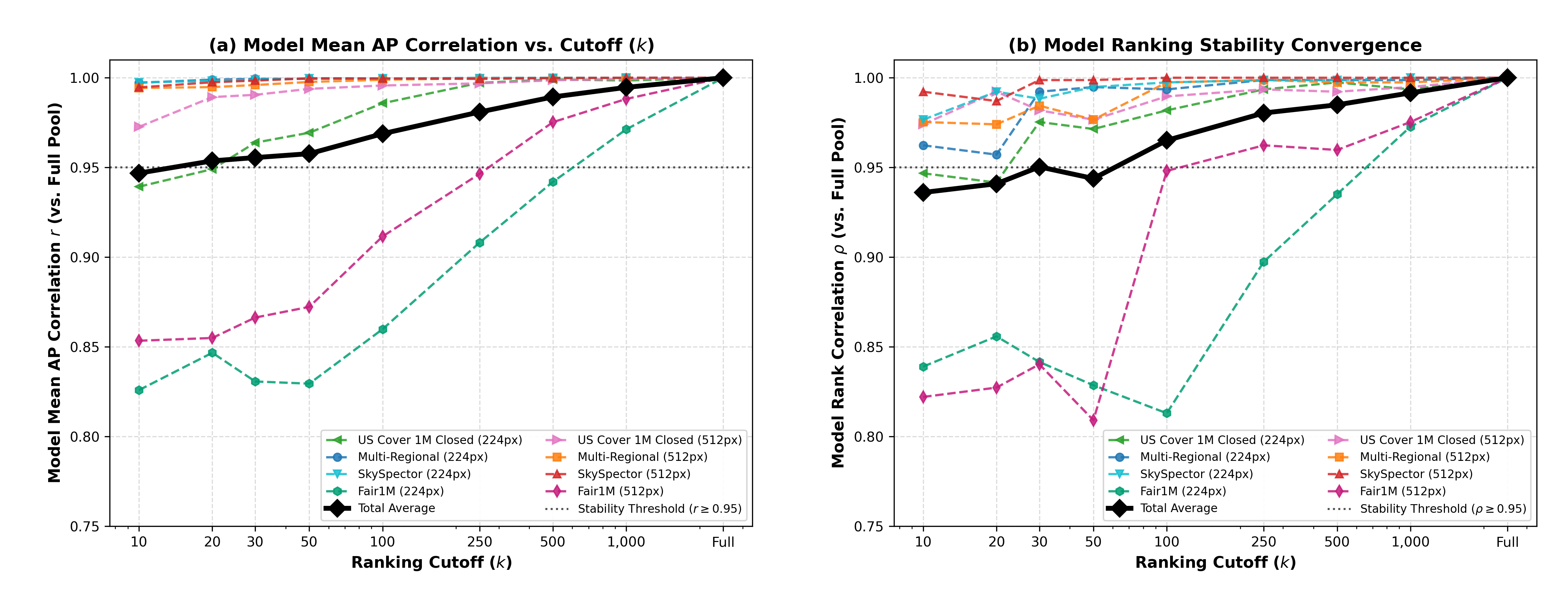}
    \caption{\textbf{Head Truncation Cutoff Analysis.} We evaluate mAP on various Multi-Retrieval benchmarks, with different cutoff top-k values. We then compute correlation for: (a) mAP metrics between cutoff values and full ranking range; (b) ranking of the models derived from their mAP between cutoff values and full ranking range. Reaching an average correlation $\rho\geq0.95$ for $k\geq30$ empirically proves that head truncation is a valid method for large-scale unlabeled datasets.}
    \label{fig:cutoff_analysis}
\end{figure}

\Cref{subsec:large_scale_dataset} presents the \textit{pooling and
bootstrapping candidate selection protocol} for making large-scale dataset annotation feasible. While this approach primarily assesses head precision within the top ranks, empirical verification confirms that relative model rankings and performance trajectories on the top-$k$ pool serve as a robust proxy for full-corpus retrieval capability. To support this claim, we show in \cref{fig:cutoff_analysis} a comparison of ranking metrics evaluated on various cutoffs, empirically proving the high correlation between top-$k$ vs. full-rank metrics and their resulting model ranks. For this experiment, 15 model variants were evaluated on various fully-labeled datasets, with different cutoffs. Correlation was calculated between the mAP scores vs. full rank (no cutoff), and between the induced model ranking vs. no cutoff. A smaller, "closed" version of the \textit{US Cover 1M} dataset was created for the sake of this experiment, constructed of smaller closed image pools for each query, containing only labeled images - this serves as a proxy for the full-scale dataset but supports calculating mAP on a full ranking.

\end{document}